\pdfoutput=1
\documentclass[11pt]{article}
\usepackage[acceptedWithA]{tacl2021v1}

\newif\ifsqueeze
\squeezefalse

\usepackage{times}
\usepackage{latexsym}
\usepackage[T1]{fontenc}
\usepackage[utf8]{inputenc}
\usepackage{microtype}
\usepackage{inconsolata}

\usepackage{amsfonts} %
\usepackage{amssymb} %
\usepackage{amsmath} %
\usepackage{mathtools}
\usepackage[capitalize,nameinlink]{cleveref} %

\usepackage[inline]{enumitem}
\ifsqueeze
\newlist{compactitem}{itemize}{3}
\setlist[compactitem]{topsep=0pt,partopsep=0pt,itemsep=0pt,parsep=0pt,leftmargin=1em}
\setlist[compactitem,1]{label=\textbullet}
\setlist[compactitem,2]{label=---}
\setlist[compactitem,3]{label=*}
\newlist{compactdesc}{description}{3}
\setlist[compactdesc]{topsep=0pt,partopsep=0pt,itemsep=0pt,parsep=0pt,leftmargin=1em}
\newlist{compactenum}{enumerate}{3}
\setlist[compactenum]{topsep=0pt,partopsep=0pt,itemsep=0pt,parsep=0pt,leftmargin=1.5em}
\setlist[compactenum,1]{label=\arabic*.}
\setlist[compactenum,2]{label=(\alph*)}
\setlist[compactenum,3]{label=\roman*.}
\else

\fi

\usepackage{comment}
\usepackage{tikz}
\usepackage{cleveref}
\usepackage{booktabs} %
\usepackage{makecell}

\usepackage{mleftright} %
\usepackage{csquotes} %

\ifsqueeze
\AtBeginDocument{%
\addtolength{\abovedisplayskip}{-4pt}%
\addtolength{\abovedisplayshortskip}{-4pt}%
\addtolength{\belowdisplayskip}{-4pt}%
\addtolength{\belowdisplayshortskip}{-4pt}%
}
\fi

\usepackage{ifthen}

\newcommand{\ignore}[1]{}

\newcommand{\DONE}[1]{}

\newcommand{\insentpara}[1]{\par\vspace{2ex}\noindent\textbf{#1}\hspace{0.33em}\ignorespaces}

\usepackage{amsthm}

\theoremstyle{definition}

\newcommand\numberthis{\addtocounter{equation}{1}\tag{\theequation}}

\usepackage[scaled=0.9]{helvet}
\usepackage{newtxmath} %

\renewcommand{\vec}[1]{\mathbf{#1}}
\newcommand{\mat}[1]{\mathbf{#1}}

\newcommand{\seq}[1]{{#1}^*}
\newcommand{\liftseq}[1]{#1}

\newcommand{\vecseq}{X}

\newcommand{\Softmax}{\mathcal{S}}
\newcommand{\argmax}{\Softmax_{\text{h}}}    \newcommand{\argmaxname}{leftmost-argmax}
\newcommand{\avgargmax}{\Softmax_{\text{a}}} \newcommand{\avgargmaxname}{average-argmax}
\newcommand{\lefthard}{leftmost-hard} \newcommand{\Lefthard}{Leftmost-hard} 
\newcommand{\righthard}{rightmost-hard}  
\newcommand{\avghard}{average-hard} \newcommand{\Avghard}{Average-hard} 
\newcommand{\soft}{softmax} \newcommand{\Soft}{Softmax}

\newcommand{\variance}{\mathrm{var}}

\newcommand{\relu}{\mathcal{R}}
\newcommand{\sigmoid}{\mathcal{\sigma}}

\newcommand{\transformerspace}{\mathbb{R}}
\newcommand{\transformerinputspace}[1]{{\seq{(\transformerspace^{#1})}}}
\newcommand{\transformerfunctionspace}{
	\transformerinputspace{d}\rightarrow \transformerinputspace{d}
}		
\newcommand{\attention}{{\mathrm{A}}}
\newcommand{\selfatt}{{\mathrm{SA}}}
\newcommand{\mha}{\mathcal{A}}

\newcommand{\layernorm}{\mathcal{N}}
\newcommand{\feedforward}{\mathcal{F}}
\newcommand{\layer}{\mathcal{L}}
\newcommand{\transformer}{\mathcal{T}}

\newcommand{\dropout}[1]{#1} %

\newcommand{\dmodel}{d}
\newcommand{\din}{{d_\text{in}}}
\newcommand{\dout}{{d_\text{out}}}
\newcommand{\dhid}{{d_\text{kv}}}
\newcommand{\dff}{{d_\text{ff}}}
\newcommand{\numheads}{H} %

\crefformat{section}{\S #2#1#3}
\crefformat{subsection}{\S #2#1#3}
\crefformat{subsubsection}{\S #2#1#3}

\newcommand{\prob}[1]{\textsc{#1}}
\newcommand{\dyck}{\prob{Dyck}}
\newcommand{\parity}{\prob{Parity}}

\newcommand{\wordsym}[1]{\ensuremath{\prob{W}(S_#1)}}

\newcommand{\probclass}[1]{\ensuremath{\mathsf{#1}}}
\newcommand{\AC}{\probclass{AC}}
\newcommand{\ACC}{\probclass{ACC}}
\newcommand{\NC}{\probclass{NC}}
\newcommand{\TC}{\probclass{TC}}

\newcommand{\nat}{\mathbb{N}_0}
\newcommand{\natpos}{\mathbb{N}}
\newcommand{\real}{\mathbb{R}}

\newcommand{\qproj}{W_\attention^\text{\kern0.5pt Q}}
\newcommand{\kproj}{W_\attention^\text{K}}
\newcommand{\vproj}{W_\attention^\text{V}}
\newcommand{\outmap}{W_\attention^\text{O}}

\newcommand{\ffone}{L_\feedforward^1}
\newcommand{\fftwo}{L_\feedforward^2}

\newcommand{\mask}{m}

\newcommand{\sym}[1]{\texttt{#1}}
\newcommand{\cls}{\texttt{CLS}}
\newcommand{\bos}{\texttt{BOS}}
\newcommand{\eos}{\texttt{EOS}}
\newcommand{\prefix}{w_{{}<t}}

\let\epsilon\varepsilon

\newcommand{\FO}{\ensuremath{\mathsf{FO}}}
\newcommand{\FOBIT}{\ensuremath{\mathsf{FO}[\mathsf{BIT}]}}
\newcommand{\FOM}{\ensuremath{\mathsf{FOM}[\mathsf{BIT}]}}

\newcommand{\logspace}{\ensuremath{\mathsf{L}}}
\newcommand{\dlogtime}{\ensuremath{\mathsf{DLOGTIME}}}
\newcommand{\poly}{\text{poly}}

\newcommand{\depth}{D} %

\title{What Formal Languages Can Transformers Express? A Survey}
\author{
         Lena Strobl \\ Ume{\aa} University, Sweden \\ \href{mailto:lena.strobl@umu.se}{\tt lena.strobl@umu.se}
         \And
         William Merrill \\ New York University, USA \\ \href{mailto:willm@nyu.edu}{\tt willm@nyu.edu}
         \And
         Gail Weiss \\ EPFL, Switzerland \\ \href{mailto:gail.weiss@epfl.ch}{\tt gail.weiss@epfl.ch}
         \AND
         David Chiang \\ University of Notre Dame, USA \\ \href{mailto:dchiang@nd.edu}{\tt dchiang@nd.edu}
         \And
         Dana Angluin\\ Yale University, USA \\ \href{mailto:dana.angluin@yale.edu}{\tt dana.angluin@yale.edu}
}

\begin{document}
\maketitle
\begin{abstract}

As transformers have gained prominence in natural language processing, some researchers have investigated theoretically what problems they can and cannot solve, by treating problems as \emph{formal languages}.
Exploring such questions can help clarify the power of transformers relative to other models of computation, their fundamental capabilities and limits, and the impact of architectural choices.
Work in this subarea has made considerable progress in recent years. Here, we undertake a comprehensive survey of this work,
documenting the diverse assumptions that underlie different results and providing a unified framework for harmonizing seemingly contradictory findings.

\end{abstract}

\section{Introduction}

Transformers \citep{vaswani-etal-2017-attention} have gained prominence in natural language processing (NLP), both in direct applications like machine translation and in pretrained models like BERT \citep{devlin2019bert} and GPT \citep{radford2018improving, gpt3, openai2023gpt4}.
Consequently, some researchers have sought to investigate their theoretical properties.
Such studies can broadly be divided into studies of \emph{expressivity} and \emph{trainability}.
While trainability is very important and the focus of much study \citep[e.g.,][]{Bhattamishra-2022-simplicity-bias,allenzhu2023physics},
here we focus on expressivity, which is a prerequisite for trainability.

Studies of expressivity could be further divided into those from the perspectives of approximation theory and of formal language theory.
The former \citep[e.g.,][]{yun-etal-2020-universal,sanford2023representational}, 
investigates transformers as approximators of various classes of \emph{functions}, along the lines of the universal approximation theorem for feedforward neural networks \citep{hornik-etal-universal-1989,cybenko:1989}.
The latter, which is the subject of this survey, investigates transformers as recognizers or generators of \emph{formal languages} -- that is, the inputs or outputs are treated as sequences of discrete symbols from a finite alphabet, and crucially as sequences of unbounded length.

The core research question in this subarea is: \emph{How can we characterize the expressivity of transformers in relation to various formal models, such as automata, boolean circuits or formal logic?}
Applications of this subarea, which are not addressed by the papers surveyed here but could be by future work, would hopefully answer questions like:
\begin{compactitem}
    \item What new transformer variants are suggested by formal models?
    \item Do failure cases anticipated from formal models occur in practice?
    \item What insights into the complexity of human language are offered by a characterization of transformer expressivity?
\end{compactitem}

This paper provides a comprehensive survey of research in this subarea.
Compared to the surveys of \citet{ackerman2020survey} and \citet{merrill2021formal,merrill-2023-black-box}, which cover convolutional neural networks (CNNs), RNNs, and transformers, this is a narrower, but deeper, survey on transformers only.

Interpreting theoretical transformer results is complex due to diverse assumptions.
Many variants of transformers exist in practice, and even more have been proposed in theory.
This diversity leads to varied, even seemingly contradictory, results.
We set up a unified framework for talking about transformer variants~(\cref{bg:transformers}), and discuss how some of these variants compare to one another in expressivity. 

We then provide background on various formal models that transformers have been compared with (\cref{sec:targets}). Then, in \cref{Se:Results}, we systematically survey current results in this literature, documenting their assumptions and claims in terms of the definitions of \cref{bg:transformers,sec:targets}.

\section{Overview}

\Cref{tab:claims} summarizes the results surveyed here. One way to classify them is into \emph{lower bounds} (what transformers \emph{can} do) and \emph{upper bounds} (what transformers \emph{can't} do).

Much work on lower bounds has looked at \emph{automata} like 
finite automata, %
counter machines, %
and Turing machines, %
all of which had been successfully related to RNNs before \citep{siegelmann+sontag:1995,merrill:2020}.
This wide diversity of machines is due to different variants of transformers, especially  whether a transformer decoder is allowed to take a number of intermediate steps before outputting a decision~(\cref{sec:cot}), which dramatically increases its power (\cref{sec:results-cot}).

By contrast, investigation of upper bounds has mainly focused on \emph{circuit complexity} (\cref{bg:circuits}), which had been successfully related to feedforward networks before \citep{parberry:1994,siu+:1995,beiu+taylor:1996,sima+orponen:2003}.
This line of research began with restricted models of transformer encoders and progressed to increasingly realistic variants and tighter bounds.
One way to restrict transformers is by discretizing the attention mechanism (\cref{sec:attention});
another is to limit the precision of number representations (\cref{sec:uniformity}).

More recent work has turned to \emph{formal logic} (\cref{bg:logic}) as a way of characterizing the expressive power of transformers.
The finer control afforded by logics opens the possibility for them to be used as upper bounds, %
lower bounds, %
or both. %

\section{Preliminaries}\label{Se:Background}

\paragraph{Sets}

We denote by $\nat=\{0,1,2,\ldots\}$ and $\natpos=\nat\setminus\{0\}$ the set of natural numbers with and without $0$, respectively.
We write $[n] = \{0,1,2,\ldots,n-1\}$ for any $n\in\natpos$. We write $\Sigma$ for a finite alphabet, which, in NLP applications, is the set of words or subwords known to the model.

\paragraph{Vectors}

We use $d$, $d'$, etc., for dimensionalities of vector spaces,
lowercase bold letters ($\vec{x}, \vec{y}, \dots$) for vectors, and uppercase bold letters ($\mat{X}, \mat{Y}, \dots$) for matrices.
For any vector $\vec{x} \in \real^d$, we number its elements starting from $0$.
For $i \in [d]$, we write $\vec{x}_i$ or $[\vec{x}]_i$ (not $x_i$) for the $i$-th component of $\vec{x}$.

\paragraph{Sequences}
For any set $A$, we write $\seq{A}$ for the set of all finite sequences over $A$.
We write the length of a sequence $s \in \seq{A}$ as $|s|$ and number its elements starting from 0;
thus, $s = s_0 s_1 \cdots s_{|s|-1}$.
We use the variable $w$ for a string in $\seq{\Sigma}$ and $n$ for the length of $w$. %
For sequences in $\seq{\real}$, we use lowercase bold letters ($\vec{x}, \vec{y}, \ldots$), and for sequences in $\transformerinputspace{d}$, we use the variable~$\vecseq$.

A function $f\colon\seq{A}\rightarrow \seq{B}$ 
is \emph{length-preserving} if 
$|f(w)|=|w|$ for all $w\in \seq{A}$.
For every function $g \colon A \rightarrow B$, we 
denote its extension to sequences by $g$ as well. That is,
$\liftseq{g}\colon\seq{A}\rightarrow\seq{B}$ is defined as follows: 
for all $s\in\seq{A}$ and $i\in[|s|]$, 
$\liftseq{g}(s)_i = g(s_i)$.

\paragraph{Neural networks}

An \emph{affine transformation} is a function~$L\colon\transformerspace^\din\rightarrow\transformerspace^\dout$ parameterized by weights $\mat{W}_L\in\transformerspace^{\dout\times \din}$ and bias $\vec{b}_L\in\transformerspace^\dout$ such that for every $\vec{x}\in\transformerspace^\din$,
$L(\vec{x}) = \mat{W}_L \vec{x} + \vec{b}_L$.
We say that $L$ is \emph{linear} if $\vec{b}_L = \vec{0}$.

The activation functions we use are the \emph{rectified linear unit} (ReLU) $\relu(x) = \max (x,0)$ and the logistic \emph{sigmoid} function $\sigmoid(x) = 1/(1+e^{-x})$.

The \emph{softmax} function $\Softmax \colon \seq{\real} \rightarrow \seq{\real}$ converts any sequence of reals into a probability distribution:
\begin{align*}
\Softmax(\vec{x})_i&=\frac{e^{\vec{x}_i}}{\sum_{i\in[|\vec{x}|]} e^{\vec{x}_i}} \qquad \forall i \in [|\vec{x}|].
\end{align*}

\section{Transformers}\label{bg:transformers}

In this section, we define transformers and relevant variants, and how transformers are used to describe formal languages.
For additional background on transformers (not in relation to formal languages), \citet{annotated-transformer} give a lucid commentary on the original paper, \citet{phuong2022formal} give formal definitions and pseudocode, and \citet{lin2022survey} survey many variants of transformers. 

Transformers are composed of an input layer (\cref{sec:embedding}), one or more hidden layers (\cref{sec:hidden}), and an output layer (\cref{sec:output}).
The inputs and outputs of the layers are sequences of vectors, which we treat as members of $\transformerinputspace{d}$.\footnote{This differs from the original paper \citep{vaswani-etal-2017-attention}, which treats them as matrices in $\transformerspace^{n \times d}$. Our notation aligns better with notation for formal languages and emphasizes the variability of the sequence length.}

\subsection{Input layer}
\label{sec:embedding}

Strings are initially mapped to sequences of vectors using a length-preserving function $e\colon\Sigma^*\rightarrow\transformerinputspace{d}$, which is the sum of a \emph{word embedding} $\text{WE} \colon \Sigma \rightarrow\transformerspace^d$ and a \emph{position(al) embedding} or \emph{encoding} $\text{PE}_n\colon [n] \rightarrow\transformerspace^d$ for $n \in \natpos$:
\begin{align*}
    e(w_0\cdots w_{n-1})_i &= \dropout{\text{WE}(w_i) + \text{PE}_n(i)}.
\end{align*}

In theoretical constructions, the word embedding can be any computable function. %

The original transformer paper \citep{vaswani-etal-2017-attention} introduced the following position embedding:
\begin{equation*}
    [\text{PE}_n(i)]_j =
    \begin{cases}
        \sin (10000^{-j/\dmodel} \cdot i) & \text{if $j$ even} \\
        \cos (10000^{-(j-1)/\dmodel} \cdot i) & \text{if $j$ odd.}
    \end{cases}
\end{equation*}
Theoretical papers have explored other position embeddings, including $i$ itself \citep{perez-etal-2021-turing}, 
$i/n$ \citep{yao-etal-2021-self,chiang-cholak-2022-overcoming}, 
and $1/i$ or $1/i^2$ \citep{perez-etal-2021-turing}.

\subsection{Hidden layers}
\label{sec:hidden}

A \emph{transformer layer} is a length-preserving function $\layer\colon\transformerfunctionspace$. 
There are two variants.
The \emph{post-norm} variant \citep{vaswani-etal-2017-attention} is
\begin{align*}
    \vecseq' &= \layernorm_1(\vecseq + \dropout{\mha(\vecseq)})\\
    \layer(\vecseq) &= \layernorm_2(\vecseq' + \dropout{\feedforward(\vecseq')}) \numberthis \label{eq:postnorm}
\end{align*}
and the \emph{pre-norm} variant \citep{wang+:2019} is
\begin{align*}
\vecseq' &= \vecseq + \dropout{\mha(\layernorm_1(\vecseq))}\\
\layer(\vecseq) &= \vecseq' + \dropout{\feedforward(\layernorm_2(\vecseq'))} \numberthis \label{eq:prenorm}
\end{align*}
where
\begin{compactitem}
    \item $\mha$ is a multi-head self-attention with $\dmodel$ input/output dimensions, $\numheads$ heads, and $\dhid$ key/value dimensions per head
    \item $\feedforward$ is a feed-forward network (\cref{sec:ffn}) with $\dmodel$ input/output dimensions and $\dff$ hidden dimensions
    \item $\layernorm_1$ and $\layernorm_2$ are layernorms with $\dmodel$ dimensions.
\end{compactitem}
We define each of these components below.

\subsubsection{Attention}
\label{sec:attention}

Attention was initially developed to facilitate retrieval of previously processed data from a variable-length history \citep{bahdanau-etal-attention}.
Transformers use a simple variant of attention known as \emph{scaled dot-product attention}.

\newcommand{\query}{\mathbf{z}}
\newcommand{\attlogit}{\mathbf{s}}
\insentpara{Scaled dot-product attention}
with $\dmodel$ input/output dimensions and $\dhid$ key/value dimensions is a function $\attention \colon \transformerspace^{\dmodel} \times \transformerinputspace{\dmodel} \rightarrow\transformerspace^{\dmodel}$ 
parameterized by linear transformations
\begin{align*}
\qproj,\kproj,\vproj\colon\transformerspace^{\dmodel}&\rightarrow\transformerspace^{\dhid} &\outmap\colon\transformerspace^{\dhid}&\rightarrow\transformerspace^{\dmodel}
\end{align*}
and defined for every $\query \in \transformerspace^{\dmodel}$,
$\vecseq\in\transformerinputspace{\dmodel}$ (with $|\vecseq| = n$),
and $j \in [n]$ as
\begin{align}
    \attlogit(\query, \vecseq)_j &= \frac{\qproj(\query) \cdot \kproj(\vecseq_j)}{\sqrt\dhid} \label{eq:att_logit}  \\
    \alpha(\query, \vecseq) &= %
    \Softmax(\attlogit(\query, \vecseq)) \label{eq:att_weight} \\
    \attention(\query, \vecseq) &= \outmap\Bigl( \sum_{j\in[n]} \alpha(\query, \vecseq)_j \, \vproj(\vecseq_j) \Bigr). \notag
\end{align}
Typically, $\attention$ is extended to a function $\attention \colon \transformerinputspace{\dmodel} \times \transformerinputspace{\dmodel} \rightarrow\transformerinputspace{\dmodel}$ 
that is length-preserving in its \emph{first} argument. In \emph{cross}-attention, $\query$ is computed by the decoder while $\vecseq$ is computed by the encoder.
In \emph{self}-attention, the two arguments are identical:
\begin{align*}
\selfatt \colon \transformerinputspace{\dmodel}&\rightarrow\transformerinputspace{\dmodel} \\
\selfatt(\vecseq) &= \attention(\vecseq, \vecseq).
\end{align*}

\paragraph{Attention masking} In \emph{future masked} (also known as \emph{causally} masked) self attention, a term $\mask(i,j)$ is added to \cref{eq:att_logit} to force every position to attend only to preceding positions:
\begin{equation*}
    \mask(i,j) =
    \begin{cases}
        0 & \text{if $j \le i$} \\
        -\infty & \text{otherwise.}
    \end{cases}
\end{equation*}
Some papers use \emph{strict} future masking, that is, $m(i,j) = 0$ iff $j<i$,
and occasionally \emph{past} masking ($j \ge i$) and strict past masking ($j>i$).

\insentpara{Multi-head attention} with $\dhid$ key/value dimensions per head is the sum of $\numheads$ attentions with $\dhid$ key/value dimensions: %
\begin{equation*}
\mha(\query,\vecseq)= \sum_{h\in[\numheads]}\attention_h(\query,\vecseq).
\end{equation*} 
Multi-head self attention is defined analogously.
This is equivalent to the original formulation, which concatenated the outputs of the heads and passed the result through a shared, larger, $\outmap$.

\paragraph{Hard attention}
Some theoretical analyses simplify attention by replacing the softmax with variants that focus attention only on the position(s) with the maximum value, breaking ties in various ways.
For any $\attlogit \in \seq\transformerspace$, let $M(\attlogit) = \{ i \in [|\attlogit|] \mid \forall j \in [|\attlogit|], \attlogit_j \le \attlogit_i \}$ be the set of indices of the maximal elements of~$\attlogit$. 
In \emph{\argmaxname}, the leftmost maximal element is used:
\begin{align*}
    [\argmax(\attlogit)]_i &= \mathbb{I}[i = \min M(\attlogit)] \\
\intertext{whereas in \emph{\avgargmaxname} the maximal elements share weight equally:}
    [\avgargmax(\attlogit)]_i &= \frac{\mathbb{I}[i \in M(\attlogit)]}{|M(\attlogit)|}.
\end{align*}
If softmax is thought of as a Boltzmann distribution, then \avgargmaxname{} is its low-temperature limit.

By substituting $\argmax$ or $\avgargmax$ for $\Softmax$ in \cref{eq:att_weight}, we get \emph{\lefthard} and \emph{\avghard} attention, respectively.
\Lefthard{} attention was previously called 
\emph{hard} attention by \citet{hahn-2020-theoretical} and
\emph{unique hard} attention by \citet{hao-etal-2022-circuits}.
One may also consider \emph{\righthard} attention, in which the rightmost maximal element is used.
\Avghard{} attention was also called 
\emph{hard} attention by \citet{perez-etal-2021-turing} and
\emph{saturated} attention by \citet{merrill-etal-2021-saturated-transformers}, and has been argued to be a realistic approximation to how trained transformers behave in practice \citep{merrill-etal-2021-effects}.

\subsubsection{Feed-forward networks}
\label{sec:ffn}

Although feed-forward networks can take many forms, in the context of transformers, we use the following definition.
A \emph{feed-forward network} (FFN) with $\dmodel$ input/output dimensions and $\dff$ hidden dimensions is a function $\feedforward \colon \transformerspace^\dmodel \rightarrow \transformerspace^\dmodel$ parameterized by two affine transformations, $\ffone \colon \transformerspace^\dmodel \to \transformerspace^\dff$ and $\fftwo \colon \transformerspace^\dff \to \transformerspace^\dmodel$, such that
\begin{align*}
    \feedforward(\vec{x}) &= \fftwo(\relu(\ffone(\vec{x})))
\end{align*}
where $\relu$ is applied component-wise.

\subsubsection{Layer normalization}
\label{sec:layernorm}

A $d$-dimensional \emph{layer normalization} \citep{ba+:2016}, or \emph{layernorm} for short, is a function $\layernorm \colon \transformerspace^d \rightarrow  \transformerspace^d$ parameterized by vectors $\gamma_\layernorm,\beta_\layernorm\in\transformerspace^d$ and scalar $\varepsilon_\layernorm\ge 0$:
\begin{equation*}
\layernorm(\vec{x}) = \gamma_\layernorm \odot \frac{ \vec{x} - \bar{\vec{x}}} {\sqrt{\variance(\vec{x})+\varepsilon_\layernorm}}+\beta_\layernorm
\end{equation*}
where $\odot$ is component-wise multiplication and
\begin{equation*}
\bar{\vec{x}} = \frac1d \sum_{i\in[d]}\vec{x}_i \qquad
\variance(\vec{x}) = \frac{1}{d}\sum_{i \in [d]}(\vec{x}_i-\bar{\vec{x}})^2.
\end{equation*}
 
The original definition of layernorm 
\citep{ba+:2016} 
sets $\varepsilon_\layernorm = 0$, but, for numerical stability, 
all implementations we are aware of set $\varepsilon_\layernorm > 0$.
Observe that $\layernorm$ is Lipschitz-continuous iff $\varepsilon_\layernorm > 0$. 

Some transformer analyses omit $\layernorm$ for simplicity \cite{perez-etal-2021-turing}, while others set $\varepsilon_\layernorm$ to achieve various effects \cite{hahn-2020-theoretical,chiang-cholak-2022-overcoming}.

\subsection{Networks and output layers}
\label{sec:output}

We now define a complete transformer network.

\subsubsection{Transformer encoders}
\label{sec:encoder}

A \emph{transformer encoder} is a length-preserving function $\transformer\colon\seq\Sigma\to\transformerinputspace{\dmodel}$
parameterized by the weights of an input layer $e$ and $\depth$ transformer layers $\layer_1,\ldots,\layer_\depth$.
A \emph{post-norm} transformer encoder is:
    \begin{equation*}
        \transformer(w) = \layer_{\depth} \circ \cdots \circ \layer_2 \circ \layer_1 \circ e(w)
    \end{equation*}
where each $\layer_l$ is a post-norm layer \labelcref{eq:postnorm}
and $\circ$ is function composition.
A \emph{pre-norm} transformer encoder is additionally parameterized by the weights of a final layernorm $\layernorm$ and is defined as:
    \begin{equation*}
        \transformer(w) = \layernorm \circ \layer_{\depth} \circ \cdots \circ \layer_2 \circ \layer_1 \circ e(w)
    \end{equation*}
where each $\layer_l$ is a pre-norm layer 
\labelcref{eq:prenorm}.

The encoder's output is a sequence of vectors in $\transformerinputspace{\dmodel}$.
To use it as a language recognizer, we add an output layer
that converts $\transformer(w)$ to a probability
\[
    \hat{p} = \sigmoid(\vec{w}\cdot[\transformer(w)]_i + b)
\]
where $\vec{w} \in \mathbb{R}^\dmodel$, $b \in \mathbb{R}$, and $i$ is a distinguished position.
The encoder accepts iff $\hat{p} \ge \frac12$. %

\Citet{chiang-cholak-2022-overcoming} also consider a requirement that an encoder accepts/rejects strings with bounded cross-entropy. That is, we say that an encoder recognizes a language $L$ with cross-entropy at most $\eta$ iff for all strings $w$, if $w \in L$ then $-\log \hat{p} \le \eta$, and if $w \not\in L$ then $-\log (1-\hat{p}) \le \eta$.

We are aware of two choices for the distinguished position $i$. Most papers use
the last position ($i = n-1$), but some \citep{chiang-cholak-2022-overcoming,chiang-2023-tighter}, inspired by binary classifiers based on BERT \citep{devlin2019bert}, prepend a special symbol \cls{} at position $0$ and use $i=0$. While this is a minor difference, it should be noted that the guarantee of exactly one occurrence of $\cls$ in the input can be useful in some constructions.

\subsubsection{Transformer decoders}
\label{sec:decoder}

A \emph{transformer decoder} is a transformer encoder $\transformer$ with future masking in its attention, typically used to generate rather than recognize strings. The input is the prefix of previously-generated symbols, $\prefix = w_0 \cdots w_{t-1}$, and the output is a probability distribution $\hat{p}(w_t \mid \prefix)$ over the next symbol,
\begin{equation*}
\hat{p}(\cdot \mid \prefix) = \Softmax(\mat{W}\,[\transformer(\prefix)]_{t-1} + \vec{b})
\end{equation*}
where $\mat{W} \in \mathbb{R}^{|\Sigma| \times \dmodel}$ and $\vec{b} \in \mathbb{R}^{|\Sigma|}$.
We assume $w_0 = \bos$ and every string ends with $\eos$, where $\bos$ and $\eos$ are special symbols that do not occur anywhere else.
To sample a string, we first sample $w_1$ from $\hat{p}(w_1 \mid \bos)$, then, for each time step $t>1$, sample $w_t$ from $\hat{p}(w_t \mid \prefix)$. The process stops when $w_t = \eos$. Because each sampled output symbol becomes part of the input at the next time step, this kind of model is called \emph{autoregressive}.

While a decoder can be used to recognize strings similarly to an encoder,
it
can also be used to generate the entire string; at least two definitions have been given for this.

First, \citet{hahn-2020-theoretical} considers a weighted language as a distribution over strings $p(w)$.
For any length~$t$, the KL divergence (relative entropy) of the model $\hat{p}(w)$ from the true distribution $p(w)$, for predicting $w_t$ conditioned on all previous words, is
    \[\Delta_t[\hat{p}\mathbin\|p] = \sum_{\prefix} \sum_{w_t} p(\prefix w_t)  \log \frac{p(w_t \mid \prefix)}{\hat{p}(w_t \mid \prefix)}.\]
As \citeauthor{hahn-2020-theoretical}'s results are negative, he does not spell out a positive criterion, but he seems to implicitly require that this divergence vanish at infinity:
\begin{equation}
\lim_{t\rightarrow\infty} \Delta_t[\hat{p}\mathbin\|p] = 0.
\label{eq:optimal_crossentropy}
\end{equation}
        
Second, let us say that a transformer decoder \emph{$\epsilon$-generates} $L$ iff
\[ L = \{ w \mid \forall t \in [|w|], \hat{p}(w_t \mid \prefix) \ge \epsilon\}.\]
Then \citet{yao-etal-2021-self}, following \citet{hewitt-etal-2020-rnns}, say that a transformer decoder $T$ generates a language $L$ iff there exists an $\epsilon > 0$ such that $T$ $\epsilon$-generates $L$.
(This means that a transformer decoder may generate more than one language, depending on the $\epsilon$ chosen.)
They also show that any $\epsilon$-generator can be converted into a recognizer.

While not focusing on transformers, \citet{lin-etal-2021-limitations} demonstrate limitations of autoregressive models for generation; for example, that there is a language $L \in \probclass{P}$ that cannot be $\epsilon$-generated in polynomial time for any $\epsilon > 0$ if $\probclass{P} \neq \probclass{NP}$.

\subsubsection{Transformer encoder--decoders}
\label{sec:encdec}

A \emph{transformer encoder--decoder} combines a transformer encoder and decoder, adding to each layer of the decoder an additional attention sublayer, known as \emph{cross attention}, which attends to the output of the encoder.
In the literature surveyed here, only the construction of \citet{perez-etal-2021-turing} and related constructions \citep{bhattamishra-2020-computational-power,wei-etal-2021-turing} employ an encoder--decoder.

\subsubsection{Intermediate steps}
\label{sec:cot}

When a transformer decoder or encoder--decoder is run as a language recognizer, it allows for the possibility of inserting a number of \emph{intermediate} time steps between the end of the input string and the decision. The encoder--decoder models above do this, as do some decoder-only models \citep{feng-etal-2023,merrill-sabharwal-2023-cot}.
As we will see (\cref{sec:results-cot}), intermediate steps vastly increase the model's power, which has also been observed in practice in the form of a ``scratchpad'' \citep{nye-2021-scratchpad} or ``chain of thought'' \citep{wei-2022-cot}.

\subsection{Uniformity and precision}
\label{sec:uniformity}

Although meaningful theoretical claims can be made about transformers for fixed-length strings \citep[e.g.][]{yun-etal-2020-universal}, it is crucial when examining transformers as language recognizers to allow for unbounded string length.
Fixing a maximum length makes all languages finite, collapsing many language classes into one.

It might be objected that considering unbounded lengths is too abstract, because in practice one can always fix a maximum length.
But this maximum length, driven by practical needs, is growing steadily: for example, GPT-4 Turbo uses 128,000 tokens of context.
At the same time, some theoretical findings surveyed here seem to have practical consequences for modest string lengths.
For example, we will see that there are reasons to think that in theory, transformers cannot recognize \prob{Parity}; in practice, they fail to learn \prob{Parity} for strings with lengths in $[2,50]$ \citep{bhattamishra-etal-2020-ability}.

Some theoretical studies of transformers do allow them to depend on the input length $n$. To borrow a term from circuit complexity (\cref{bg:circuits}), they allow certain kinds of \emph{non-uniformity}. As we have seen, some position embeddings (\cref{sec:embedding}) depend on $n$.
We discuss some other instances below.

\paragraph{Numeric precision}

Transformers operate, in principle, on real numbers.
While hard attention transformers could be defined using only rational numbers, even rational numbers can represent an arbitrary amount of information.
With RNNs, the use of real or rational numbers has led to results that make them appear more powerful in theory than in practice \citep{siegelmann+sontag:1994,siegelmann+sontag:1995,weiss-etal-2018-practical}.

Consequently, many studies use limited-precision numbers. 
Some studies limit number representations to have $O(1)$ bits, as floating-point numbers do in practice \citep{chiang-2023-tighter}. %
But \citet{merrill-2023-majority} argue that in $O(1)$ precision, attention cannot attend uniformly to a string of sufficient length $n$, as the attention weights ($\alpha$) would all round down to zero.
So $O(\log n)$ bits of precision is a common choice \citep{yao-etal-2021-self,merrill-2022-log,merrill-2023-majority}.
Other choices are possible as well: \citet{merrill-2022-log} use the set $\mathbb{F} = \{a/2^b \mid a \in \mathbb{Z}, b \in \natpos\}$.

Restricting intermediate activations to limited precision introduces many decisions about when and how rounding should take place, which can potentially affect expressivity.
For example, when summing $n$ numbers, one could round after each addition or only at the end of the summation.
Better formalizing these decisions and their impact on expressivity is an area for future research.

\paragraph{Parameters}

A few constructions allow the parameters themselves to depend on $n$, which we consider to be a stronger dependence, because if these transformers were to be learned from data, different transformers would have to be learned for different maximum lengths. 
Finally, a few papers construct transformers in which $d$, and therefore the number of parameters, depends on $n$, which we consider to be stronger still.

\subsection{Summary} \label{sec:categories}
In summary, transformers can vary in at least the following ways, any of which could \emph{a priori} impact theoretical claims:
\begin{compactitem}
\item Architecture: encoder-only, decoder-only, or encoder--decoder
\item For encoders: definition of recognition
\item For decoders and encoder--decoders: definition of generation and how many intermediate steps 
\item Position embedding (PE)
\item Attention pattern: \lefthard, \righthard, \avghard, or \soft
\item Attention masking: none, future, or past
\item Layernorm: inclusion or omission, value of $\varepsilon_\layernorm$
\item Residual connections: pre-norm or post-norm
\item Precision: infinite, $O(\log n)$, $O(1)$
\item Uniformity: whether parameter values or number of parameters depend on $n$.
\end{compactitem}

\section{Languages and Language Classes}
\label{sec:targets}

Next, we present various formal models that transformers are compared to in the literature surveyed.

\subsection{Automata and classes $\logspace$, $\probclass{NL}$, $\probclass{P}$}
\label{bg:automata}

We assume familiarity with finite automata and Turing machines; for definitions, please see the textbook by \citet{sipser-2013-introduction}. 
Counter machines are automata with integer-valued registers \citep{fischer+:1968}; they have been studied extensively in connection with LSTM RNNs \citep{weiss-etal-2018-practical,suzgun-etal-2019-lstm,merrill-2019-sequential,merrill:2020}. 

The language classes $\logspace$ 
(languages decidable in $O(\log n)$ space) and $\probclass{P}$ (languages decidable in polynomial time) are defined using deterministic Turing machines (with a read-only input tape and a read/write work tape). 
The class \probclass{NL} (languages decidable in nondeterministic $O(\log n)$ space) uses nondeterministic Turing machines.
The class \dlogtime{} (languages decidable in $O(\log n)$ time) uses random-access Turing machines~\citep{Barrington1988OnUW}.
It is known that \[ \probclass{L} \subseteq \probclass{NL} \subseteq \probclass{P} \] but none of these inclusions are known to be strict.

\subsection{Circuits and classes $\AC^0$, $\ACC^0$, $\TC^0$, $\NC^1$}
\label{bg:circuits}

Circuits are a model of parallel computation particularly relevant to transformers. For more details, please see the textbook by \citet{arora-barak-computational-complexity}.

Circuits operate on binary values. If we choose a fixed-length encoding of the symbols of $\Sigma$ as strings of $b = \lceil \log_2 |\Sigma| \rceil$ bits, then a circuit can simulate input alphabet $\Sigma$ by encoding the value of the $i$-th input symbol into positions $ib$ to $ib+(b-1)$. For the rest of this section, we assume $\Sigma = \{0,1\}$.

\paragraph{Circuits}
A \emph{circuit} $C$ with input length $n$ is a directed acyclic graph with $n$ \emph{input} vertices $s_1, \ldots, s_n$ and zero or more \emph{gate} vertices, each labeled with a \emph{type} NOT, AND, or OR.
Input vertices have fan-in (in-degree) zero, NOT gates have fan-in one, and the fan-in of AND and OR gates can be either two or unbounded.
One (input or gate) vertex $t$ is designated the \emph{output} of the circuit.

Given an input string $w \in \{0,1\}^n$, each input vertex $s_i$ is assigned the value $w_i$, and each gate vertex is assigned the value computed by applying the logical function corresponding to its type to the values assigned to its in-neighbors.
The circuit computes the boolean function $C \colon \{0,1\}^n \rightarrow \{0,1\}$, mapping each input string to the value assigned to $t$.
The \emph{depth} of $C$, denoted $\depth(C)$, is the length of the longest directed path from any $s_i$ to $t$.
The \emph{size} of $C$, denoted $|C|$, is the number of vertices in $C$.

\paragraph{Circuit families}
A \emph{circuit family} is a sequence $\mathcal{C} = \{C_n\}_{n\in\natpos}$ such that for each $n$, $C_n$ is a circuit with input length $n$.
We treat $\mathcal{C}$ as a function on $\{0,1\}^*$ as follows:
for every $w\in \{0,1\}^*$, $\mathcal{C}(w) = C_{|w|}(w)$.
Then $\mathcal{C}$ defines the language $L(\mathcal{C}) = \{w\in \{0,1\}^* \mid \mathcal{C}(w)=1 \}$, and we say that $\mathcal{C}$ recognizes $L(\mathcal{C})$.
The \emph{depth} and \emph{size} of $\mathcal{C}$ are the functions $n \mapsto \depth(C_n)$ and $n \mapsto |C_n|$.

\paragraph{Uniformity}
As defined, a circuit family contains a different circuit for each length $n$, with no constraint on the relationship between the circuits.
For example, let $L$ be any \emph{unary} language: $L \subseteq \{1\}^*$.
For $n \in\natpos$, if $1^n \not\in L$, define $C_n$ to be a circuit for the constant $0$ function (an OR gate with fan-in $0$), and if $1^n \in L$, define $C_n$ to be a circuit for the AND of all the inputs.
Thus, every unary language, even an undecidable one, is recognized by a circuit family of size $O(n)$ and depth $O(1)$.

A uniformity restriction on a circuit family $\{C_n\}_{n\in\natpos}$ requires that the task of constructing a description of the circuit $C_n$ given input $n$ be computable within some specified resource bound as a function of $n$, potentially making it comparable with classes defined by bounds on Turing machine time or space.
Two such uniformity bounds are used in the work here: \probclass{L} and \dlogtime.
Because these bounds are very restrictive, a special representation of the circuit $C_n$ is used, namely, the ability to answer queries of the type of a gate and whether the output of one gate is an input to another gate.

We assume that the vertices of the circuit $C_n$ are numbered from $0$ to $|C_n|-1$.
The \emph{direct connection language} of a family of circuits $\cal{C}$ is the set of all tuples $\langle f, i, j, \texttt{1}^n\rangle$ such that in $C_n$, vertex $i$ has type $f$ 
and there is an edge from vertex $i$ to vertex $j$ \citep{Barrington1988OnUW}.
Given a computable function bounding the size of $\cal{C}$ 
and access to a membership oracle for the direct connection language, for any $n$ it is straightforward to write out the list of vertices, edges, and types in~$C_n$.

Then a circuit family $\cal{C}$ is \emph{\logspace-uniform} (resp., \emph{\dlogtime-uniform}) if there is a Turing machine that runs in logarithmic space (resp., deterministic logarithmic time) to decide membership in the direct connection language of~$\cal{C}$.

\paragraph{Circuit complexity classes} 
Circuit complexity classes classify circuit families and the languages they recognize based on uniformity, depth, size, fan-in bound, and the allowed gates.
Since transformers have constant depth, circuit classes with constant depth are of particular interest; the classes that are used in the work we survey are:
\begin{compactitem}
\item $\AC^0$ contains those languages that can be recognized by families of circuits with unbounded fan-in, constant depth, and polynomial size.
\item $\ACC^0$ is like $\AC^0$, but also has gates that output 1 iff the inputs sum to $0$ modulo some constant.
\item $\TC^0$ is like $\AC^0$, but also allows MAJORITY gates, which have unbounded fan-in and output $1$ iff at least half of their inputs are $1$.
\item $\NC^1$ is like $\AC^0$, but with fan-in at most 2 and depth in $O(\log n)$.
\end{compactitem}
The known relationships between these classes are:
\[\AC^0 \subsetneq \ACC^0 \subseteq \TC^0 \subseteq \NC^1\]
in the \dlogtime-uniform, \logspace-uniform, and non-uniform settings;
moreover, \logspace-uniform $\NC^1 \subseteq \logspace$.

\subsection{Logic}\label{bg:logic}

A formal language can also be defined as a set of finite strings that satisfy a closed formula of a logic.
For more details, refer to \citet{Thomas1997} or \citet{straubing:1994}.

In the \emph{first-order logic of strings}, or $\FO$, the formulas are the smallest set containing:
\begin{compactitem}
\item Variables $x, y$, and so on.
\item Atomic formulas $Q_a(x)$, $x=y$, $x < y$, where $a \in \Sigma$ is a symbol and $x, y$ are variables.
\item $\phi_1 \land \phi_2$, $\phi_1 \lor \phi_2$, $\phi_1 \rightarrow \phi_2$, $\neg \phi_1$, where $\phi_1$ and $\phi_2$ are formulas.
\item $\forall x. \phi$, $\exists x. \phi$, where $x$ is a variable and $\phi$ is a formula.
\end{compactitem}
Under the intended interpretation, variables stand for positions of a finite string $w$, and $Q_a(x)$ is true iff $w_x = a$.
For example, if $\Sigma=\{a,b\}$, $\forall x. \forall y. Q_a(x) \land Q_b(y) \rightarrow x < y$ defines the regular language $a^\ast b^\ast$. The language defined by a closed formula $\phi$ consists of those strings that satisfy $\phi$.

The languages definable in $\FO$ are exactly the \emph{star-free} languages \citep{mcnaughton+papert:1971}. %
Other variants add more quantifiers:
\begin{compactitem}
\item $\mathsf{FOC}$ adds counting quantifiers $\exists^{=x}y.\phi$, which hold iff there are exactly $x$ values of $y$ that make $\phi$ true \citep{Barrington1988OnUW}.
\item $\mathsf{FOM}$ adds majority quantifiers $\mathsf{M}x. \phi$, which hold iff at least half of the values of $x$ make $\phi$ true \citep{Barrington1988OnUW}.
\end{compactitem}
We are also interested in various sets of predicates:
\begin{compactitem}
\item Modular predicates $\mathsf{MOD}^r_m(x)$, which hold iff $x \equiv r \pmod{m}$ \citep{barrington-etal-1992-rl-nc1}.
\item $\mathsf{BIT}(x,y)$, which holds iff the $y$-th bit of $x$ is~$1$.
\item $\mathsf{Mon}$, the set of all predicates on one position, possibly depending on $n$.\footnote{Although \citet{mixbarrington-etal-2005} define $\textsf{Mon}$ to be the collection of all monadic predicates without dependence on $n$, \citet{barcelo-etal-2023} do allow them to depend on $n$.}
\item $\mathsf{ARB}$, the set of all predicates on one or more positions.
\end{compactitem}
A logic extended with predicates is conventionally written with the predicates in square brackets; for example, we write $\FOBIT$ for first-order logic with the $\mathsf{BIT}$ predicate.

In \emph{linear temporal logic} or LTL \citep{kamp:1968}, every formula implicitly depends on a single time (or position).
There are atomic formulas $Q_a$ for every $a \in \Sigma$, the connectives $\land$, $\lor$, and $\neg$, as well as operators $\textbf{since}$ and $\textbf{until}$. The formula $\alpha \mathrel{\textbf{since}} \beta$ is true iff $\alpha$ was true at some past time $i$ and $\beta$ was true from $i$ to now (exclusive). LTL is equivalent to $\FO$ \citep{kamp:1968}.

\subsection{Relationships}
\label{sec:classes}

\begin{figure}
    \centering
    \includegraphics[width=\linewidth]{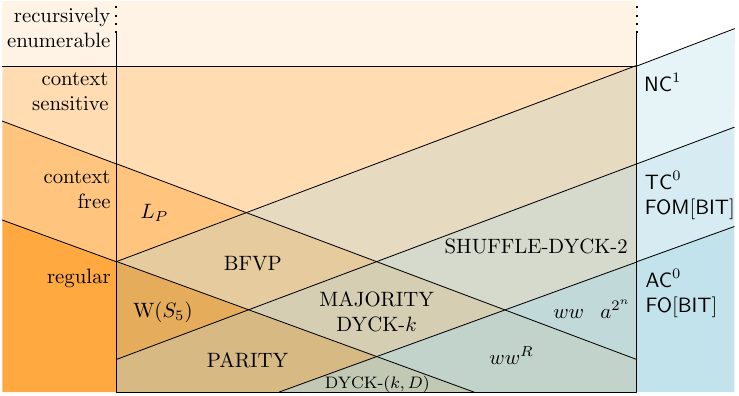}
    \caption{Relationship of some languages and language classes discussed in this paper (right) to the Chomsky hierarchy (left), 
    assuming that $\TC^0 \subsetneq \NC^1$ and $\logspace \subsetneq \probclass{NL}$.
    Circuit classes are \dlogtime-uniform.
    } \label{fig:classes}
\end{figure}

\Cref{fig:classes}, which depicts the relationships between the language classes defined above, shows that the classes defined by circuits/logics cut across the (perhaps more familiar) Chomsky hierarchy. In this figure and in this section, all circuit classes are understood to be \dlogtime-uniform unless specified otherwise.

\subsubsection{Beyond $\AC^0$} 

The classic examples of languages not in $\AC^0$ are $\prob{Parity}$ and $\prob{Majority}$.
The language $\prob{Parity} \subseteq \{\sym{0},\sym{1}\}^*$ contains all bit strings containing an odd number of~$\sym{1}$'s, and
$\prob{Majority} \subseteq \{\sym{0},\sym{1}\}^*$ consists of all bit strings in which more than half of the bits are~$\sym{1}$'s.
Other problems in $\TC^0$ but not $\AC^0$ include 
sorting, integer multiplication \citep{chandra+:1984}, and integer division \citep{hesse:2001}.

\paragraph{Dyck languages}

{
\renewcommand{\lparen}[1]{\texttt{(}_{#1}}
\renewcommand{\rparen}[1]{\texttt{)}_{#1}}
The language $\prob{Dyck-}k$ for $k>0$ is the language of strings over $k$ pairs of parentheses that are correctly balanced and nested.
If we write the $i$-th parenthesis pair as $\lparen{i}\;\rparen{i}$ for each $i \in [k]$, then $\prob{Dyck-}k$ is generated by the context-free grammar $\{ S \rightarrow \lparen{i} S \rparen{i} S \mid i \in [k] \} \cup \{ S \rightarrow \epsilon \}$.
These languages are of interest because any context-free language can be obtained by applying a string homomorphism to the intersection of a Dyck language with a regular language \citep{chomsky+schutzenberger:1963}. 

Some papers surveyed here consider variations on Dyck languages.
The language \prob{Dyck-$(k,D)$} for $D>0$ is the subset of \prob{Dyck-$k$} consisting of strings with maximum nesting depth~$D$; it is a star-free regular language (and therefore in $\AC^0$).

The language \prob{Shuffle-Dyck-$k$} is the set of strings over $k$ pairs of parentheses in which, for each parenthesis pair, erasing the other types of parentheses leaves a correctly balanced and nested string.  
For example, $[(()])$ is in \prob{Shuffle-Dyck-2}.
If $k > 1$, \prob{Shuffle-Dyck-$k$} is not context free.

\subsubsection{Beyond $\TC^0$}

As we will see (\cref{sec:results-tc0}), some transformer variants lie within %
$\TC^0$. What problems lie beyond?

\paragraph{The word problem for permutation groups}

A permutation of $[k]$ is a bijection $\pi \colon [k] \rightarrow [k]$, and $S_k$ is the set of all permutations of $[k]$.
Treating $S_k$ as an alphabet and compositions of permutations as strings, we can define the language $\wordsym{k}$ of compositions of permutations of $[k]$ that equal the identity permutation.
For example, in $S_3$, the permutation $(120)$ maps $0 \mapsto 1$, $1 \mapsto 2$, and $2 \mapsto 0$, so that $\wordsym3$ contains $(120)\circ(120)\circ(120)$ but not $(120)\circ(120)$.
These languages are easy for finite automata to recognize, but difficult with only fixed computation depth.
Indeed, $\wordsym5$ is complete for 
$\NC^1$ under 
$\AC^0$ reductions \citep{barrington:1989}, so it is not in 
$\TC^0$, assuming that $\TC^0 \subsetneq \NC^1$ (as is widely believed). This makes it an example of a regular language that transformer encoders probably cannot recognize.

The languages $\wordsym{k}$ have some relevance to natural language: they resemble expressions like \emph{the child of the enemy of Ann} where the interpretation of \emph{the child of} is (roughly) a permutation of possible referents \citep{paperno-2022-learning},
and problems that have been used to benchmark transformers' state-tracking abilities \citep{kim-schuster-2023-entity}.

\insentpara{Other languages} that are widely believed to be not in
$\TC^0$ include:
\begin{compactitem}
\item The language of closed Boolean formulas that are true (\prob{BFVP}) is context-free but complete for 
$\NC^1$ under \dlogtime{} reductions \citep{buss-1987}, so it is outside %
$\TC^0$ if $\TC^0 \subsetneq \NC^1$.
\item Undirected graph connectivity is $\logspace$-complete under \logspace-uniform $\NC^1$ reductions \citep{cook-mckenzie-1987-logspace,reingold2008undirected}, so it is outside \logspace-uniform $\NC^1$ (and therefore outside $\TC^0$) if $\text{\logspace-uniform $\NC^1$} \subsetneq \logspace$.
\item There is a context-free language $L_P$ that is $\probclass{NL}$-complete under $\logspace$ reductions \citep{sudborough-1975}, so it is outside \logspace{} (and therefore outside 
$\NC^1$ and $\TC^0$) if $\logspace \subsetneq \mathsf{NL}$.
\item Solving systems of linear equalities and universal context-free grammar recognition are $\mathsf{P}$-complete under \logspace{} reductions \citep{jones-1976-polynomial,greenlaw-etal-1995}, so they are outside 
$\TC^0$ if $\logspace \subsetneq \mathsf{P}$.
\item Matrix permanent is %
known to be outside of 
$\TC^0$ \citep{allender:1999}.
\end{compactitem}

\subsubsection{Circuits and logics}

\dlogtime-uniform $\AC^0$ and $\TC^0$ are equivalent to $\FOBIT$ and $\FOM$, respectively. There are many such equivalences between circuit classes and logics. As a rule of thumb, 
adding unbounded fan-in gates to a circuit family correlates with adding quantifiers to the corresponding logic, and
increasing the degree of non-uniformity of a circuit family correlates with adding numerical predicates to the corresponding logic \citep{barrington-immerman-1994}.
For example, making $\AC^0$ and $\TC^0$ completely non-uniform corresponds to adding arbitrary numerical predicates $(\mathsf{ARB})$ to $\FO$ and $\mathsf{FOM}$, respectively \citep{immerman:1997,Barrington1988OnUW}.

As we will see below, circuits and logics have their advantages and disadvantages for capturing the expressivity of transformers.
An advantage of the circuit approach is that they have a more transparent resemblance to transformers.
Transformers are computations with bounded depth, so it's not hard to see that they should be computable by circuit families with bounded depth ($\AC^0$ or $\TC^0$).
On the other hand, an advantage of the logical approach is that if we seek an exact characterization of transformers, it can be easier in a logic to add or remove quantifiers or predicates, to limit quantifier depth or number of variables, to partition terms into different sorts, and so on, than to make adjustments to a circuit family.

\section{Current Results}
\label{Se:Results}

\newcommand{\mycite}[2]{\hyperlink{cite.#2}{#1}~\citeyear{#2}} %

\begin{table*}\centering\resizebox{\linewidth}{!}{%
\begin{tabular}{@{}lllll@{}}
\toprule
Lower bound & Source & PE & Attention & Notes  \\
\midrule
$\ni \prob{Majority}$ & \citealt{perez2019turing} & none & \avghard & \\ %
$\ni \prob{Shuffle-Dyck-}k$ & \citealt{bhattamishra-etal-2020-ability} & none & \soft, future mask & \\ %
$\supseteq \text{SSCMs}$ & \citealt{bhattamishra-etal-2020-ability} & none & \soft, future mask & \\ %
$\ni \text{\prob{Dyck}-k}$ & \citealt{yao-etal-2021-self} & $i/n$, $i/n^3$, $n$ & \soft{} \& \lefthard{}  & \\ %
$\supseteq \probclass{P}$ & \citealt{perez-etal-2021-turing} & $i, 1/i, 1/{i^2}$ & \avghard{} & $\poly(n)$ steps \\ %
$\ni \prob{Parity}$ & \citealt{chiang-cholak-2022-overcoming} & $i/n$, $(-1)^i$ & \soft & \\ %
$\supseteq \probclass{FOC[MOD;\mathord+]}$ & \citealt{chiang-2023-tighter} & sinusoidal & \soft &  \\ %
$\supseteq \FO[\textsf{Mon}]$ & \citealt{barcelo-etal-2023} & arbitrary & \lefthard \\ %
$\supseteq \textsf{LTL+C}[\textsf{Mon}]$ & \citealt{barcelo-etal-2023} & arbitrary & \avghard \\ %
\midrule[\heavyrulewidth]
Upper bound & Source & Precision & Attention & Notes \\
\midrule
$\not\ni \prob{Parity}, \text{\prob{Dyck}-1}$ & \citealt{hahn-2020-theoretical} & $\real$ & \lefthard \\
$\not\ni \prob{Parity}, \text{\prob{Dyck}-2}$ & \citealt{hahn-2020-theoretical} & $\real$ & \soft, future mask & $\varepsilon_\layernorm>0$, vanishing KL \\
$\subseteq \AC^0$ & \citealt{hao-etal-2022-circuits} & $\mathbb{Q}$ & \lefthard &  \\ %
$\subseteq \TC^0$ & \citealt{merrill-etal-2021-saturated-transformers} & $\mathbb{F}$ & \avghard \\
$\subseteq \probclass{FOC[MOD;\mathord+]}$ & \citealt{chiang-2023-tighter} & $O(1)$ & \soft &  \\
$\subseteq \text{\probclass{L}-uniform~$\TC^0$}$ & \mycite{Merrill \& Sabharwal}{merrill-2022-log} & $O(\log n)$ & \soft \\
$\subseteq \FOM$ & 
\mycite{Merrill \& Sabharwal}{merrill-2023-majority} & $O(\log n)$ & \soft \\
$\subseteq \text{\probclass{L}-uniform~$\TC^0$}$ &  \citealt{strobl2023averagehard} & $\mathbb{F}$ & \avghard \\
\midrule[\heavyrulewidth]
Equivalent & Source & PE & Attention & Notes \\
\midrule
$= \probclass{RE}$ & \citealt{perez-etal-2021-turing} & $i, 1/i, 1/{i^2}$ & \avghard{} & unbounded steps \\ %
$= \FO$ & \citealt{angluin-etal-2023} & none & \righthard, strict future mask \\ %
$= \FO[\textsf{MOD}]$ & \citealt{angluin-etal-2023} & sinusoidal & \righthard, strict future mask \\
$= \FO[\textsf{Mon}]$ & \citealt{angluin-etal-2023} & arbitrary & \righthard, strict future mask \\
$= \probclass{P}$ & \mycite{Merrill \& Sabharwal}{merrill-sabharwal-2023-cot} & none & \avghard, future mask & $\poly(n)$ steps \\ %
\bottomrule
\end{tabular}%
}
\caption{Surveyed claims and their assumptions. Please see the main text for full details of assumptions.}
\label{tab:claims}
\end{table*}

While this area of research still has many unresolved questions, the emerging picture has three levels of expressivity. At the upper end are decoders or encoder--decoders with intermediate steps; these are equivalent to Turing machines (\cref{sec:results-cot}). 
At the lower end are encoders with \lefthard{} or \righthard{} attention; these can recognize only languages in $\AC^0$ (\cref{sec:results-hard}).
In the middle are encoders  with \avghard{} or \soft{} attention, which are the least well-understood but appear to lie between $\AC^0$ and $\TC^0$  (\cref{sec:results-soft}).

In this section, ``transformer'' refers to a transformer encoder unless otherwise indicated.

\subsection{Decoders with intermediate steps}
\label{sec:results-cot}

\Citet{perez-etal-2021-turing} consider transformer encoder--decoders with several modifications:
\begin{compactitem}
\item The PE includes components $i$, $1/i$, and $1/i^2$.
\item In self attention, \cref{eq:att_logit} takes the negative absolute value of the dot-product, and \cref{eq:att_weight} uses \avghard{} attention.
\item The FFNs use sigmoids instead of ReLUs.
\end{compactitem}
As described above (\cref{sec:encdec}), the decoder is allowed to run for arbitrarily many time steps until an acceptance criterion is met.
Under these assumptions, transformer encoder--decoders can recognize any recursively enumerable language.\footnote{\Citet{perez-etal-2021-turing} 
define both Turing machines and encoder--decoders to halt only when accepting. %
The construction could easily be modified to capture decidable languages.}
This result uses arbitrary precision, but as a corollary, they show that a $T(n)$-time-bounded Turing machine can be simulated in a transformer using $O(\log T(n))$ precision and $O(T(n))$ intermediate steps.

\Citet{bhattamishra-2020-computational-power} provide a simpler proof of \citeauthor{perez-etal-2021-turing}'s result by reducing to an RNN and appealing to the construction of \citet{siegelmann+sontag:1995}. They do this for two sets of assumptions. First, 
\begin{compactitem}
\item The PE includes only $i$.
\item The self attention sublayers are as above.
\item The FFNs use saturated linear activation functions: $\sigma(x) = \max (0, \min (1, x))$.
\end{compactitem}
Second, they show the same with no PE and standard dot-product attention with future masking.

\Citet{wei-etal-2021-turing} define a notion of \emph{statistically-meaningful} (SM) approximation and show that transformer encoder--decoders SM-approximate Turing machines.
Both the decoder and Turing machine are limited to $N$ time steps; additionally,
\begin{compactitem}
\item The PE can be an arbitrary computable function on $[N]$.
\item Attention is \avghard. %
\item The FFNs have three ReLU layers.
\end{compactitem}

\Citet{feng-etal-2023} observe that the problems of evaluating arithmetic expressions or solving linear equations over $\mathbb{Z}_p$ are $\NC^1$-hard under $\dlogtime$ reductions, so (if $\TC^0 \subsetneq \NC^1$) they cannot be solved by $O(\log n)$-precision transformer decoders without intermediate steps.\footnote{This uses the result of \citet{merrill-2023-majority}, which would have to be adapted to transformer decoders, but this should be straightforward.} Similarly, the universal recognition problem for CFGs is $\probclass{P}$-complete, so (if $\logspace \subsetneq \probclass{P}$) it cannot be solved by $O(\log n)$-precision transformer decoders without intermediate steps.

However, these problems can be solved by a transformer decoder using (a polynomial number of) intermediate steps.
The decoder has GELU activations \citep{hendrycks-gimpel-2016} and PE including $i$ and (for linear equation solving) $m^2 \sin \frac{2i\pi}{m}$ and $m^2 \cos\frac{2i\pi}{m}$ where $m$ is the number of variables. More generally, they define a class of dynamic-programming algorithms that these transformers can solve using intermediate steps. All these decoders have parameters that depend on $n$.

\Citet{merrill-sabharwal-2023-cot} show that a transformer decoder with $O(\log(n+T(n)))$ precision and $O(T(n))$ intermediate steps can simulate a Turing machine for $T(n)$ steps, and in particular, decoders with a polynomial number of intermediate steps recognize \emph{exactly} the languages in $\probclass{P}$. The proof is similar to that of 
\citet{perez-etal-2021-turing}, but uses a standard definition of transformers without PEs, relying only on the mild assumption that the input string begins with $\bos$.

\subsection{\Lefthard/\righthard{} attention}
\label{sec:results-hard}

\Citet{hahn-2020-theoretical} shows that \lefthard{} attention transformers cannot recognize \parity{} or \dyck-1, using a variant of \citeauthor{furst+:1984}'s random restriction method for proving that \parity{} is outside of $\AC^0$.

\citet{hao-etal-2022-circuits} show more generally that any language recognized by a transformer with \lefthard{} attention is in $\AC^0$.
The proof gives a normal form for transformers with \lefthard{} attention and uses it to construct an $\AC^0$ circuit family.
It uses the fact that only $O(\log n)$ bits of information are needed per position.

\Citet{barcelo-etal-2023} give a lower bound on \lefthard-attention transformers with arbitrary PEs depending on a single position $i$ and length $n$, including $i$, $\tfrac{1}{i+1}$, $(-1)^i$, $\cos \frac{\pi(1-2^{-i})}{10}$, and $\sin \frac{\pi(1-2^{-i})}{10}$. They show that these transformers can recognize any language definable in $\FO[\textsf{Mon}]$. Their proof converts a $\FO[\textsf{Mon}]$ formula to LTL (\cref{bg:logic}), which is simulated in a transformer.

\Citet{angluin-etal-2023} exactly characterize \righthard-attention transformers with strict future masking. 
Without PEs, these transformers recognize exactly the class of star-free languages, that is, languages definable in $\FO$. 
With periodic PEs,
they are exactly equivalent to $\FO[\mathsf{MOD}]$, and with arbitrary PEs, they are exactly equivalent to $\FO[\mathsf{Mon}]$. Strict masking is important, as nonstrict masking is less expressive. They give two proofs of the star-free to transformer direction, one which goes through LTL (\cref{bg:logic}) and one which uses Krohn-Rhodes theory. These proofs use a Boolean-valued version of RASP \citep{weiss-etal-2021-rasp} as an intermediate representation.

\subsection{\Avghard{} and \soft{} attention}
\label{sec:results-soft}

Theoretical results on \avghard{} and \soft{} attention transformers have not yet clearly separated the two, so we treat them together. 
Both kinds of attention enable counting, which can be used to solve problems like \prob{Majority} that are outside $\AC^0$.
But these transformers are no more powerful than \dlogtime-uniform $\TC^0$, implying that they likely cannot solve problems complete for $\NC^1$, $\mathsf L$, and other classes believed to be above $\TC^0$ (\cref{sec:classes}).

\subsubsection{Lower bounds: particular languages}
\label{sec:results-langs}

The languages \prob{Majority}, \prob{Dyck-$k$}, and \prob{Parity}  are all not in $\AC^0$, so are interesting test cases.

\Citet{perez2019turing} prove that a transformer encoder--decoder with a trivial decoder and without any PE recognizes \prob{Majority}; \citet{merrill-etal-2021-saturated-transformers} prove the same for transformer encoders. 

\Citet{bhattamishra-etal-2020-ability} prove that $\prob{Shuffle-Dyck-}k$ (which equals $\prob{Dyck-}1$ when $k = 1$) is recognizable by a soft-attention transformer with future masking, no PE, no layernorm, and no residual connections.
\Citet{yao-etal-2021-self} show that a transformer decoder can generate \prob{Dyck}-$k$ using $O(\log n)$ precision, \soft{} and \lefthard{} attention, future masking, and a PE including $i/n$, $i/n^3$, and~$n$. They also give constructions for \prob{Dyck-$(k,D)$}.

\Citet{chiang-cholak-2022-overcoming} show that transformers whose PE includes $i/n$ and $(-1)^i = \cos i \pi$ can recognize \parity. 

On the other hand, \citet{hahn-2020-theoretical} shows that \soft{} attention transformers cannot generate \parity{} or \dyck-2 under the following two conditions:
\begin{compactenum}
\item\label{item:lipschitz} all position-wise functions are Lipschitz-continuous, and
\item\label{item:entropy} generation is defined using the KL divergence criterion in \cref{eq:optimal_crossentropy}. 
\end{compactenum}

The apparent contradiction is resolved by considering the different assumptions underlying each result. 
\Citet{chiang-cholak-2022-overcoming} address this by giving two constructions corresponding to Hahn's two conditions. The first has Lipschitz-continuous position-wise functions, but has high cross-entropy (\cref{sec:encoder}); as a generator, it would not meet criterion~\labelcref{eq:optimal_crossentropy}. The second construction uses layernorm with $\varepsilon_\layernorm = 0$, which is not Lipschitz-continuous, but it has arbitrarily low cross-entropy.

A number of authors have tested empirically whether transformers can learn the above languages.
\Citet{ebrahimi-etal-2020-self} find that they are competitive with LSTMs at learning \prob{Dyck}-$2$ and \mbox{\prob{Dyck}-$4$}, and that prepending a \bos{} symbol helps.

\Citet{bhattamishra-etal-2020-ability} train transformers with future masking and no PE on 
\prob{Dyck-$1$} and
\prob{Shuffle-Dyck-$k$}, finding near-perfect learning and length generalization.
For the languages \prob{Dyck-$(1,D)$} with learned or sinusoidal PEs, 
they find that the models do not generalize well for $D>1$.
\citet{yao-etal-2021-self} then investigate \prob{Dyck-$(k,D)$} for several values of $k$ and $D$ and several PEs.
They report strong generalization only when using $i/n$ for the PE, 
and posit that this is the key.
It is hard, however, to directly compare the two results:
\citet{bhattamishra-etal-2020-ability} require correct prediction of the possible next symbols at each string prefix,
while \citet{yao-etal-2021-self} average over predictions of right brackets.

\Citet{deletang-etal-2023-chomsky} study experimentally how well transformers (and other networks) learn tasks at various levels of the Chomsky hierarchy, including generalization to longer strings.
They find that transformers learn \prob{Majority}, but not $\prob{Parity}$.

\subsubsection{Upper bounds: $\TC^0$}
\label{sec:results-tc0}

\Citet{merrill-etal-2021-saturated-transformers} prove an upper bound analogous to that of \citet{hao-etal-2022-circuits}, but for \avghard-attention transformers.
They show that an \avghard-attention transformer with activations in $\mathbb{F}$ can be simulated in $\TC^0$.
\Citet{strobl2023averagehard} tightens this bound to \logspace-uniform $\TC^0$.

Furthermore, \citet{merrill-2022-log} show that \soft{} attention, $O(\log n)$-precision transformers are in \logspace-uniform $\TC^0$, and then tighten this bound to \dlogtime-uniform $\TC^0$ \citep{merrill-2023-majority}.
The proof
constructs subroutines to answer queries about the types of nodes and connectivity of pairs of nodes in the computation graph of a transformer, and shows that these queries can be translated to queries for a $\TC^0$ circuit family with $O(\log n)$ time overhead.

An upper bound of \dlogtime-uniform $\TC^0$ immediately implies an upper bound of $\FOM$ \citep{merrill-2023-majority}. 
\Citet{chiang-2023-tighter} prove a tighter upper bound using a logic called $\probclass{FOC[MOD;\mathord+]}$, but on transformers with $O(1)$ precision. This result is discussed below.

\subsubsection{Other lower bounds} 
\label{sec:results-other}

In addition to explicit constructions for particular languages mentioned above, various lower bounds have been proven, which are quite diverse.

\paragraph{Counter machines} \Citet{bhattamishra-etal-2020-ability}, following \citet{merrill-etal-2020-formal}, define a subclass of counter machines called \emph{simplified and stateless $k$-counter machines} (SSCMs). 
These %
can update each counter based on the current input symbol, but have no state and cannot read the counters until the end of the string.
They show that any SSCM can be converted to an equivalent transformer with future masking and no residual connections.

\paragraph{Finite automata} \Citet{Liu-2022-shortcuts} study the ability of transformers with future masked attention to simulate deterministic finite automata (DFAs), in the sense of computing not only the same acceptance decision but also the same state sequence.
Although a transformer with depth $N$ can simulate a DFA for $N$ timesteps, \citeauthor{Liu-2022-shortcuts}~show how to construct lower-depth \emph{shortcuts} for subclasses roughly corresponding to classes of regular languages in \cref{fig:classes}.
Though the parameters of these constructions depend on $N$,
in the context of this survey, 
a noteworthy finding is that any regular language in $\mathsf{ACC}^0$ can be recognized up to length $N$ by a transformer whose FFNs use sine activations and whose \emph{number} of parameters is independent of $N$. 

\paragraph{First-order logic} \Citet{chiang-2023-tighter} obtain both an upper and a lower bound by  defining a logic $\probclass{FOC[MOD;\mathord+]}$, which is first-order logic with counting quantifiers, using two sorts for positions and counts \citep[p.~185--187]{immerman:1999}, where positions have the $\mathsf{MOD}$ predicate (but not $<$ or $=$), and counts have $<$, $+$, and $=$, capturing the fact that transformers can add and compare activations, but not positions. They show that this logic is intermediate in expressivity between $O(1)$-precision and infinite-precision transformers.
The lower-bound proof uses
a normal form that eliminates quantifiers over counts and makes quantifiers over positions have depth~1;
a perhaps surprising consequence is that $O(1)$-precision transformers are no more powerful than 2-layer uniform-attention transformers.

\paragraph{Temporal logic} \Citet{barcelo-etal-2023} show that \avghard-attention transformers with arbitrary PEs depending on a single position $i$ and length $n$, including $i$, $\tfrac{1}{i+1}$, $(-1)^i$, $\cos \frac{\pi(1-2^{-i})}{10}$, and $\sin \frac{\pi(1-2^{-i})}{10}$, can recognize any language definable in LTL with counting operators, Presburger arithmetic on counts, and predicates in $\mathsf{Mon}$.

\paragraph{Programming languages} \citet{weiss-etal-2021-rasp} introduce the RASP (Restricted Access Sequence Processing) language as an abstraction of transformers, discussing how its components relate to the transformer architecture.
However, they do not prove any relationship.

\Citet{lindner+:2023} present Tracr, a compiler from RASP programs to transformers. To do so, they impose some restrictions: a maximum input length, given at compile time; a mandatory $\bos$ token; and the removal of \emph{selector composition}, a RASP operation with no clear parallel in transformers. 
They rewrite several programs from \citet{weiss-etal-2021-rasp} without this operation.
In the other direction, \citet{friedman2023learning} define 
a restricted class of transformers that
can be learned and decompiled into RASP.
Finally, \citet{angluin-etal-2023} use a version of RASP restricted to Boolean values, and \citet{zhou-etal-2023-rasp-length-generalization} use a restricted version of RASP to explore length generalization.

\section{Conclusions}

Out of the large body of research surveyed above, we highlight several conclusions:
\begin{compactenum}
    \item Transformer decoders can use intermediate steps to simulate Turing machines; with unbounded steps, they are Turing-complete.
    \item Regarding the expressivity of transformer encoders, circuit complexity and logic are especially promising frameworks.
    \item \Lefthard-attention transformer encoders are in $\AC^0$ and cannot solve some intuitively easy problems, like \prob{Parity} and \prob{Majority}.
    \item \Soft{} and \avghard{} attention give transformer encoders the ability to count. Still, they lie within $\TC^0$ and likely cannot solve problems like evaluating closed Boolean formulas.
\end{compactenum}
Some open questions that we think should be priorities for future research are:
\begin{compactenum}[resume]
    \item Some variants (PEs, average-hard vs.~softmax attention, pre-norm vs.~post-norm, the presence of \bos/\eos/\cls) appear to be instrumental in proofs reviewed here; can their effect on expressivity be clarified?
    \item Can the expressivity of \soft-attention transformers be characterized more tightly or even exactly in terms of some logic?
    \item Given the current practical importance of decoder-only transformers and chain-of-thought, what further insights can circuits or logic provide into transformer decoders?
\end{compactenum}
We hope this paper can serve as a valuable resource for researchers pursuing these and other questions.

\section*{Acknowledgements}

We would like to thank 
Frank Drewes,
Jon Rawski,
Ashish Sabharwal, and 
the anonymous reviewers
for their valuable comments on earlier versions of this paper.
      
\bibliography{bib}

\providecommand{\noopsort}[1]{}
\begin{thebibliography}{89}
\expandafter\ifx\csname natexlab\endcsname\relax\def\natexlab#1{#1}\fi

\bibitem[{Ackerman and Cybenko(2020)}]{ackerman2020survey}
Joshua Ackerman and George Cybenko. 2020.
\newblock \href {http://arxiv.org/abs/2006.01338} {A survey of neural networks
  and formal languages}.
\newblock {arXiv}:2006.01338.

\bibitem[{Allen-Zhu and Li(2023)}]{allenzhu2023physics}
Zeyuan Allen-Zhu and Yuanzhi Li. 2023.
\newblock \href {https://arxiv.org/abs/2305.13673} {Physics of language models:
  Part 1, context-free grammar}.
\newblock {arXiv}:2305.13673.

\bibitem[{Allender(1999)}]{allender:1999}
Eric Allender. 1999.
\newblock \href {http://cjtcs.cs.uchicago.edu/articles/1999/7/contents.html}
  {The permanent requires large uniform threshold circuits}.
\newblock \emph{Chicago Journal of Theoretical Computer Science}, 1999(7).

\bibitem[{Angluin et~al.(2023)Angluin, Chiang, and Yang}]{angluin-etal-2023}
Dana Angluin, David Chiang, and Andy Yang. 2023.
\newblock \href {https://arxiv.org/abs/2310.13897} {Masked hard-attention
  transformers and {B}oolean {RASP} recognize exactly the star-free languages}.
\newblock {arXiv}:2310.13897.

\bibitem[{Arora and Barak(2009)}]{arora-barak-computational-complexity}
Sanjeev Arora and Boaz Barak. 2009.
\newblock \href
  {http://www.cambridge.org/catalogue/catalogue.asp?isbn=9780521424264}
  {\emph{Computational Complexity: {A} Modern Approach}}.
\newblock Cambridge University Press.

\bibitem[{Ba et~al.(2016)Ba, Kiros, and Hinton}]{ba+:2016}
Jimmy~Lei Ba, Jamie~Ryan Kiros, and Geoffrey~E. Hinton. 2016.
\newblock \href {https://arxiv.org/abs/1607.06450} {Layer normalization}.
\newblock In \emph{NIPS 2016 Deep Learning Symposium}.

\bibitem[{Bahdanau et~al.(2015)Bahdanau, Cho, and
  Bengio}]{bahdanau-etal-attention}
Dzmitry Bahdanau, Kyunghyun Cho, and Yoshua Bengio. 2015.
\newblock \href {http://arxiv.org/abs/1409.0473} {Neural machine translation by
  jointly learning to align and translate}.
\newblock In \emph{Proceedings of the Third International Conference on
  Learning Representations (ICLR)}.

\bibitem[{Barcel{\'o} et~al.(2024)Barcel{\'o}, Kozachinskiy, Lin, and
  Podolskii}]{barcelo-etal-2023}
Pablo Barcel{\'o}, Alexander Kozachinskiy, Anthony~Widjaja Lin, and Vladimir
  Podolskii. 2024.
\newblock \href {https://openreview.net/forum?id=gbrHZq07mq} {Logical languages
  accepted by transformer encoders with hard attention}.
\newblock In \emph{Proceedings of the Twelfth International Conference on
  Learning Representations (ICLR)}.

\bibitem[{Barrington(1989)}]{barrington:1989}
David~A. Barrington. 1989.
\newblock \href {https://doi.org/10.1016/0022-0000(89)90037-8} {Bounded-width
  polynomial-size branching programs recognize exactly those languages in
  $\mathit{NC^1}$}.
\newblock \emph{Journal of Computer and System Sciences}, 38(1):150--164.

\bibitem[{Barrington et~al.(1992)Barrington, Compton, Straubing, and
  Thérien}]{barrington-etal-1992-rl-nc1}
David~A. Barrington, Kevin Compton, Howard Straubing, and Denis Thérien. 1992.
\newblock \href {https://doi.org/https://doi.org/10.1016/0022-0000(92)90014-A}
  {Regular languages in $\mathit{NC^1}$}.
\newblock \emph{Journal of Computer and System Sciences}, 44(3):478--499.

\bibitem[{Barrington et~al.(2005)Barrington, Immerman, Lautemann, Schweikardt,
  and Thérien}]{mixbarrington-etal-2005}
David A.~Mix Barrington, Neil Immerman, Clemens Lautemann, Nicole Schweikardt,
  and Denis Thérien. 2005.
\newblock \href {https://doi.org/10.1016/j.jcss.2004.07.004} {First-order
  expressibility of languages with neutral letters or: The {C}rane {B}each
  conjecture}.
\newblock \emph{Journal of Computer and System Sciences}, 70(2):101--127.

\bibitem[{Barrington et~al.(1990)Barrington, Immerman, and
  Straubing}]{Barrington1988OnUW}
David A.~Mix Barrington, Neil Immerman, and Howard Straubing. 1990.
\newblock \href {https://doi.org/https://doi.org/10.1016/0022-0000(90)90022-D}
  {On uniformity within $\mathit{NC^1}$}.
\newblock \emph{Journal of Computer and System Sciences}, 41(3):274--306.

\bibitem[{Barrington and Immerman(1994)}]{barrington-immerman-1994}
David~Mix Barrington and Neil Immerman. 1994.
\newblock \href {https://doi.org/10.1109/SCT.1994.315806} {Time, hardware, and
  uniformity}.
\newblock In \emph{Proceedings of the IEEE 9th Annual Conference on Structure
  in Complexity Theory}, pages 176--185.

\bibitem[{Beiu and Taylor(1996)}]{beiu+taylor:1996}
Valeriu Beiu and John~G. Taylor. 1996.
\newblock \href {https://doi.org/10.1016/0893-6080(96)00130-X} {On the circuit
  complexity of sigmoid feedforward neural networks}.
\newblock \emph{Neural Networks}, 9(7):1155--1171.

\bibitem[{Bhattamishra et~al.(2020{\natexlab{a}})Bhattamishra, Ahuja, and
  Goyal}]{bhattamishra-etal-2020-ability}
Satwik Bhattamishra, Kabir Ahuja, and Navin Goyal. 2020{\natexlab{a}}.
\newblock \href {https://doi.org/10.18653/v1/2020.emnlp-main.576} {On the
  ability and limitations of {T}ransformers to recognize formal languages}.
\newblock In \emph{Proceedings of the 2020 Conference on Empirical Methods in
  Natural Language Processing (EMNLP)}, pages 7096--7116.

\bibitem[{Bhattamishra et~al.(2020{\natexlab{b}})Bhattamishra, Patel, and
  Goyal}]{bhattamishra-2020-computational-power}
Satwik Bhattamishra, Arkil Patel, and Navin Goyal. 2020{\natexlab{b}}.
\newblock \href {https://doi.org/10.18653/v1/2020.conll-1.37} {On the
  computational power of {T}ransformers and its implications in sequence
  modeling}.
\newblock In \emph{Proceedings of the 24th Conference on Computational Natural
  Language Learning (CoNLL)}, pages 455--475.

\bibitem[{Bhattamishra et~al.(2023)Bhattamishra, Patel, Kanade, and
  Blunsom}]{Bhattamishra-2022-simplicity-bias}
Satwik Bhattamishra, Arkil Patel, Varun Kanade, and Phil Blunsom. 2023.
\newblock \href {https://doi.org/10.18653/v1/2023.acl-long.317} {Simplicity
  bias in {T}ransformers and their ability to learn sparse {B}oolean
  functions}.
\newblock In \emph{Proceedings of the 61st Annual Meeting of the Association
  for Computational Linguistics (ACL)}, pages 5767--5791.

\bibitem[{Brown et~al.(2020)Brown, Mann, Ryder, Subbiah, Kaplan, Dhariwal,
  Neelakantan, Shyam, Sastry, Askell, Agarwal, Herbert-Voss, Krueger, Henighan,
  Child, Ramesh, Ziegler, Wu, Winter, Hesse, Chen, Sigler, Litwin, Gray, Chess,
  Clark, Berner, McCandlish, Radford, Sutskever, and Amodei}]{gpt3}
Tom~B. Brown, Benjamin Mann, Nick Ryder, Melanie Subbiah, Jared Kaplan,
  Prafulla Dhariwal, Arvind Neelakantan, Pranav Shyam, Girish Sastry, Amanda
  Askell, Sandhini Agarwal, Ariel Herbert-Voss, Gretchen Krueger, Tom Henighan,
  Rewon Child, Aditya Ramesh, Daniel~M. Ziegler, Jeffrey Wu, Clemens Winter,
  Christopher Hesse, Mark Chen, Eric Sigler, Mateusz Litwin, Scott Gray,
  Benjamin Chess, Jack Clark, Christopher Berner, Sam McCandlish, Alec Radford,
  Ilya Sutskever, and Dario Amodei. 2020.
\newblock \href
  {https://proceedings.neurips.cc/paper_files/paper/2020/file/1457c0d6bfcb4967418bfb8ac142f64a-Paper.pdf}
  {Language models are few-shot learners}.
\newblock In \emph{Advances in Neural Information Processing Systems 33
  (NeurIPS)}, pages 1877--1901.

\bibitem[{Buss(1987)}]{buss-1987}
Samuel~R. Buss. 1987.
\newblock \href {https://doi.org/10.1145/28395.28409} {The {B}oolean formula
  value problem is in {ALOGTIME}}.
\newblock In \emph{Proceedings of the Nineteenth Annual ACM Symposium on Theory
  of Computing (STOC)}, pages 123--131.

\bibitem[{Chandra et~al.(1984)Chandra, Stockmeyer, and Vishkin}]{chandra+:1984}
Ashok~K. Chandra, Larry Stockmeyer, and Uzi Vishkin. 1984.
\newblock \href {https://doi.org/10.1137/0213028} {Constant depth
  reducibility}.
\newblock \emph{SIAM J. Computing}, 13(2):423--439.

\bibitem[{Chiang and Cholak(2022)}]{chiang-cholak-2022-overcoming}
David Chiang and Peter Cholak. 2022.
\newblock \href {https://doi.org/10.18653/v1/2022.acl-long.527} {Overcoming a
  theoretical limitation of self-attention}.
\newblock In \emph{Proceedings of the 60th Annual Meeting of the Association
  for Computational Linguistics (ACL)}, pages 7654--7664.

\bibitem[{Chiang et~al.(2023)Chiang, Cholak, and Pillay}]{chiang-2023-tighter}
David Chiang, Peter Cholak, and Anand Pillay. 2023.
\newblock \href {https://proceedings.mlr.press/v202/chiang23a.html} {Tighter
  bounds on the expressivity of transformer encoders}.
\newblock In \emph{Proceedings of the 40th International Conference on Machine
  Learning (ICML)}, volume 202 of \emph{Proceedings of Machine Learning
  Research}, pages 5544--5562.

\bibitem[{Chomsky and Schützenberger(1963)}]{chomsky+schutzenberger:1963}
N.~Chomsky and M.~P. Schützenberger. 1963.
\newblock \href {https://doi.org/10.1016/S0049-237X(08)72023-8} {The algebraic
  theory of context-free languages}.
\newblock In P.~Braffort and D.~Hirschberg, editors, \emph{Computer Programming
  and Formal Systems}, volume~35 of \emph{Studies in Logic and the Foundations
  of Mathematics}, pages 118--161. Elsevier.

\bibitem[{Cook and McKenzie(1987)}]{cook-mckenzie-1987-logspace}
Stephen~A. Cook and Pierre McKenzie. 1987.
\newblock \href {https://doi.org/10.1016/0196-6774(87)90018-6} {Problems
  complete for deterministic logarithmic space}.
\newblock \emph{Journal of Algorithms}, 8(3):385--394.

\bibitem[{Cybenko(1989)}]{cybenko:1989}
G.~Cybenko. 1989.
\newblock \href {https://doi.org/10.1007/BF02551274} {Approximation by
  superpositions of a sigmoidal function}.
\newblock \emph{Mathematics of Control, Signals, and Systems}, 2(4):303--314.

\bibitem[{Del{\'e}tang et~al.(2023)Del{\'e}tang, Ruoss, Grau-Moya, Genewein,
  Wenliang, Catt, Cundy, Hutter, Legg, Veness, and
  Ortega}]{deletang-etal-2023-chomsky}
Gr{\'e}goire Del{\'e}tang, Anian Ruoss, Jordi Grau-Moya, Tim Genewein, Li~Kevin
  Wenliang, Elliot Catt, Chris Cundy, Marcus Hutter, Shane Legg, Joel Veness,
  and Pedro~A. Ortega. 2023.
\newblock \href {https://openreview.net/forum?id=WbxHAzkeQcn} {Neural networks
  and the {C}homsky hierarchy}.
\newblock In \emph{Proceedings of the Eleventh International Conference on
  Learning Representations (ICLR)}.

\bibitem[{Devlin et~al.(2019)Devlin, Chang, Lee, and
  Toutanova}]{devlin2019bert}
Jacob Devlin, Ming-Wei Chang, Kenton Lee, and Kristina Toutanova. 2019.
\newblock \href {https://aclanthology.org/N19-1423} {{BERT}: Pre-training of
  deep bidirectional {T}ransformers for language understanding}.
\newblock In \emph{Proceedings of the 2019 Conference of the North American
  Chapter of the Association for Computational Linguistics: Human Language
  Technologies (NAACL HLT)}, pages 4171--4186.

\bibitem[{Ebrahimi et~al.(2020)Ebrahimi, Gelda, and
  Zhang}]{ebrahimi-etal-2020-self}
Javid Ebrahimi, Dhruv Gelda, and Wei Zhang. 2020.
\newblock \href {https://doi.org/10.18653/v1/2020.findings-emnlp.384} {How can
  self-attention networks recognize {D}yck-n languages?}
\newblock In \emph{Findings of the Association for Computational Linguistics:
  EMNLP 2020}, pages 4301--4306.

\bibitem[{Feng et~al.(2023)Feng, Zhang, Gu, Ye, He, and Wang}]{feng-etal-2023}
Guhao Feng, Bohang Zhang, Yuntian Gu, Haotian Ye, Di~He, and Liwei Wang. 2023.
\newblock \href
  {https://papers.nips.cc/paper_files/paper/2023/hash/dfc310e81992d2e4cedc09ac47eff13e-Abstract-Conference.html}
  {Towards revealing the mystery behind {C}hain of {T}hought: A theoretical
  perspective}.
\newblock In \emph{Advances in Neural Information Processing Systems 36
  (NeurIPS)}, pages 70757--70798.

\bibitem[{Fischer et~al.(1968)Fischer, Meyer, and Rosenberg}]{fischer+:1968}
Patrick~C. Fischer, Albert~R. Meyer, and Arnold~L. Rosenberg. 1968.
\newblock \href {https://doi.org/10.1007/BF01694011} {Counter machines and
  counter languages}.
\newblock \emph{Mathematical Systems Theory}, 2:265--283.

\bibitem[{Friedman et~al.(2023)Friedman, Wettig, and
  Chen}]{friedman2023learning}
Dan Friedman, Alexander Wettig, and Danqi Chen. 2023.
\newblock \href
  {https://papers.nips.cc/paper_files/paper/2023/hash/995f693b73050f90977ed2828202645c-Abstract-Conference.html}
  {Learning {T}ransformer programs}.
\newblock In \emph{Advances in Neural Information Processing Systems 36
  (NeurIPS)}, pages 49044--49067.

\bibitem[{Furst et~al.(1984)Furst, Saxe, and Sipser}]{furst+:1984}
Merrick Furst, James~B. Saxe, and Michael Sipser. 1984.
\newblock \href {https://doi.org/10.1007/BF01744431} {Parity, circuits, and the
  polynomial-time hierarchy}.
\newblock \emph{Mathematical Systems Theory}, 17:13--27.

\bibitem[{Greenlaw et~al.(1995)Greenlaw, Hoover, and
  Ruzzo}]{greenlaw-etal-1995}
Raymond Greenlaw, H.~James Hoover, and Walter~L. Ruzzo. 1995.
\newblock \emph{Limits to Parallel Computation: {P}-Completeness Theory}.
\newblock Oxford University Press.
\newblock Preliminary version of Appendix~A available as
  \href{https://doi.org/10.7939/R39Z90F7X}{Technical Report TR91-11, University
  of Alberta, Department of Computing Science}.

\bibitem[{Hahn(2020)}]{hahn-2020-theoretical}
Michael Hahn. 2020.
\newblock \href {https://doi.org/10.1162/tacl_a_00306} {Theoretical limitations
  of self-attention in neural sequence models}.
\newblock \emph{Transactions of the Association for Computational Linguistics},
  8:156--171.

\bibitem[{Hao et~al.(2022)Hao, Angluin, and Frank}]{hao-etal-2022-circuits}
Yiding Hao, Dana Angluin, and Robert Frank. 2022.
\newblock \href {https://doi.org/10.1162/tacl_a_00490} {Formal language
  recognition by hard attention {T}ransformers: Perspectives from circuit
  complexity}.
\newblock \emph{Transactions of the Association for Computational Linguistics},
  10:800--810.

\bibitem[{Hendrycks and Gimpel(2016)}]{hendrycks-gimpel-2016}
Dan Hendrycks and Kevin Gimpel. 2016.
\newblock \href {https://arxiv.org/abs/1606.08415} {{G}aussian error linear
  units ({GELU}s)}.
\newblock {arXiv}:1606.08415.

\bibitem[{Hesse(2001)}]{hesse:2001}
William Hesse. 2001.
\newblock \href {https://doi.org/10.1007/3-540-48224-5_9} {Division is in
  uniform {TC$^0$}}.
\newblock In \emph{Automata, Languages and Programming (ICALP)}, pages
  104--114. Springer.

\bibitem[{Hewitt et~al.(2020)Hewitt, Hahn, Ganguli, Liang, and
  Manning}]{hewitt-etal-2020-rnns}
John Hewitt, Michael Hahn, Surya Ganguli, Percy Liang, and Christopher~D.
  Manning. 2020.
\newblock \href {https://doi.org/10.18653/v1/2020.emnlp-main.156} {{RNN}s can
  generate bounded hierarchical languages with optimal memory}.
\newblock In \emph{Proceedings of the 2020 Conference on Empirical Methods in
  Natural Language Processing (EMNLP)}, pages 1978--2010.

\bibitem[{Hornik et~al.(1989)Hornik, Stinchcombe, and
  White}]{hornik-etal-universal-1989}
Kurt Hornik, Maxwell~B. Stinchcombe, and Halbert White. 1989.
\newblock \href {https://doi.org/10.1016/0893-6080(89)90020-8} {Multilayer
  feedforward networks are universal approximators}.
\newblock \emph{Neural Networks}, 2(5):359--366.

\bibitem[{Huang et~al.(2022)Huang, Subramanian, Sum, Almubarak, and
  Biderman}]{annotated-transformer}
Austin Huang, Suraj Subramanian, Jonathan Sum, Khalid Almubarak, and Stella
  Biderman. 2022.
\newblock \href {http://harvardnlp.github.io/annotated-transformer} {The
  annotated {T}ransformer}.
\newblock Based on original version by Sasha Rush.

\bibitem[{Immerman(1997)}]{immerman:1997}
Neil Immerman. 1997.
\newblock \href {https://doi.org/10.1137/0216051} {Languages that capture
  complexity classes}.
\newblock \emph{SIAM Journal on Computing}, 16(4):760--778.

\bibitem[{Immerman(1999)}]{immerman:1999}
Neil Immerman. 1999.
\newblock \emph{Descriptive Complexity}.
\newblock Springer.

\bibitem[{Jones and Laaser(1976)}]{jones-1976-polynomial}
Neil~D. Jones and William~T. Laaser. 1976.
\newblock \href {https://doi.org/10.1016/0304-3975(76)90068-2} {Complete
  problems for deterministic polynomial time}.
\newblock \emph{Theoretical Computer Science}, 3(1):105--117.

\bibitem[{Kamp(1968)}]{kamp:1968}
Johan Anthony~Willem Kamp. 1968.
\newblock \href {https://www.proquest.com/docview/302320357} {\emph{Tense Logic
  and the Theory of Linear Order}}.
\newblock Ph.D. thesis, University of California, Los Angeles.

\bibitem[{Kim and Schuster(2023)}]{kim-schuster-2023-entity}
Najoung Kim and Sebastian Schuster. 2023.
\newblock \href {https://doi.org/10.18653/v1/2023.acl-long.213} {Entity
  tracking in language models}.
\newblock In \emph{Proceedings of the 61st Annual Meeting of the Association
  for Computational Linguistics (Volume 1: Long Papers)}, pages 3835--3855.

\bibitem[{Lin et~al.(2021)Lin, Jaech, Li, Gormley, and
  Eisner}]{lin-etal-2021-limitations}
Chu-Cheng Lin, Aaron Jaech, Xin Li, Matthew~R. Gormley, and Jason Eisner. 2021.
\newblock \href {https://doi.org/10.18653/v1/2021.naacl-main.405} {Limitations
  of autoregressive models and their alternatives}.
\newblock In \emph{Proceedings of the 2021 Conference of the North American
  Chapter of the Association for Computational Linguistics: Human Language
  Technologies (NAACL HLT)}, pages 5147--5173.

\bibitem[{Lin et~al.(2022)Lin, Wang, Liu, and Qiu}]{lin2022survey}
Tianyang Lin, Yuxin Wang, Xiangyang Liu, and Xipeng Qiu. 2022.
\newblock \href {https://doi.org/10.1016/j.aiopen.2022.10.001} {A survey of
  transformers}.
\newblock \emph{AI Open}, 3:111--132.

\bibitem[{Lindner et~al.(2023)Lindner, Kram{\'a}r, Rahtz, McGrath, and
  Mikulik}]{lindner+:2023}
David Lindner, J{\'a}nos Kram{\'a}r, Matthew Rahtz, Thomas McGrath, and
  Vladimir Mikulik. 2023.
\newblock \href
  {https://papers.nips.cc/paper_files/paper/2023/hash/771155abaae744e08576f1f3b4b7ac0d-Abstract-Conference.html}
  {{T}racr: Compiled transformers as a laboratory for interpretability}.
\newblock In \emph{Advances in Neural Information Processing Systems 36
  (NeurIPS)}, pages 37876--37899.

\bibitem[{Liu et~al.(2023)Liu, Ash, Goel, Krishnamurthy, and
  Zhang}]{Liu-2022-shortcuts}
Bingbin Liu, Jordan~T. Ash, Surbhi Goel, Akshay Krishnamurthy, and Cyril Zhang.
  2023.
\newblock \href {https://openreview.net/forum?id=De4FYqjFueZ} {Transformers
  learn shortcuts to automata}.
\newblock In \emph{Proceedings of the Eleventh International Conference on
  Learning Representations (ICLR)}.

\bibitem[{McNaughton and Papert(1971)}]{mcnaughton+papert:1971}
Robert McNaughton and Seymour~A. Papert. 1971.
\newblock \href {https://archive.org/details/CounterFre_00_McNa}
  {\emph{Counter-Free Automata}}.
\newblock MIT Press.

\bibitem[{Merrill(2019)}]{merrill-2019-sequential}
William Merrill. 2019.
\newblock \href {https://doi.org/10.18653/v1/W19-3901} {Sequential neural
  networks as automata}.
\newblock In \emph{Proceedings of the Workshop on Deep Learning and Formal
  Languages: Building Bridges}, pages 1--13.

\bibitem[{Merrill(2020)}]{merrill:2020}
William Merrill. 2020.
\newblock \href {https://arxiv.org/abs/2004.06866} {On the linguistic capacity
  of real-time counter automata}.
\newblock {arXiv}:2004.06866.

\bibitem[{Merrill(2021)}]{merrill2021formal}
William Merrill. 2021.
\newblock \href {https://arxiv.org/abs/2102.10094} {Formal language theory
  meets modern {NLP}}.
\newblock {arXiv}:2102.10094.

\bibitem[{Merrill(2023)}]{merrill-2023-black-box}
William Merrill. 2023.
\newblock \href {https://doi.org/10.1007/978-3-031-33264-7_1} {Formal languages
  and the {NLP} black box}.
\newblock In \emph{Developments in Language Theory}, pages 1--8.

\bibitem[{Merrill et~al.(2021)Merrill, Ramanujan, Goldberg, Schwartz, and
  Smith}]{merrill-etal-2021-effects}
William Merrill, Vivek Ramanujan, Yoav Goldberg, Roy Schwartz, and Noah~A.
  Smith. 2021.
\newblock \href {https://doi.org/10.18653/v1/2021.emnlp-main.133} {Effects of
  parameter norm growth during transformer training: Inductive bias from
  gradient descent}.
\newblock In \emph{Proceedings of the 2021 Conference on Empirical Methods in
  Natural Language Processing (EMNLP)}, pages 1766--1781.

\bibitem[{Merrill and Sabharwal(2023{\natexlab{a}})}]{merrill-2022-log}
William Merrill and Ashish Sabharwal. 2023{\natexlab{a}}.
\newblock \href
  {https://direct.mit.edu/tacl/article/doi/10.1162/tacl_a_00562/116413/The-Parallelism-Tradeoff-Limitations-of-Log}
  {\noopsort{a}{T}he parallelism tradeoff: Limitations of log-precision
  transformers}.
\newblock \emph{Transactions of the Association for Computational Linguistics},
  11:531--545.

\bibitem[{Merrill and Sabharwal(2023{\natexlab{b}})}]{merrill-2023-majority}
William Merrill and Ashish Sabharwal. 2023{\natexlab{b}}.
\newblock \href
  {https://papers.nips.cc/paper_files/paper/2023/hash/a48e5877c7bf86a513950ab23b360498-Abstract-Conference.html}
  {\noopsort{b}{A} logic for expressing log-precision transformers}.
\newblock In \emph{Advances in Neural Information Processing Systems 36
  (NeurIPS)}, pages 52453--52463.

\bibitem[{Merrill and Sabharwal(2024)}]{merrill-sabharwal-2023-cot}
William Merrill and Ashish Sabharwal. 2024.
\newblock \href {https://openreview.net/forum?id=NjNGlPh8Wh} {\noopsort{c}{T}he
  expressive power of transformers with chain of thought}.
\newblock In \emph{Proceedings of the Twelfth International Conference on
  Learning Representations (ICLR)}.

\bibitem[{Merrill et~al.(2022)Merrill, Sabharwal, and
  Smith}]{merrill-etal-2021-saturated-transformers}
William Merrill, Ashish Sabharwal, and Noah~A. Smith. 2022.
\newblock \href {https://doi.org/10.1162/tacl_a_00493} {Saturated transformers
  are constant-depth threshold circuits}.
\newblock \emph{Transactions of the Association for Computational Linguistics},
  10:843--856.

\bibitem[{Merrill et~al.(2020)Merrill, Weiss, Goldberg, Schwartz, Smith, and
  Yahav}]{merrill-etal-2020-formal}
William Merrill, Gail Weiss, Yoav Goldberg, Roy Schwartz, Noah~A. Smith, and
  Eran Yahav. 2020.
\newblock \href {https://doi.org/10.18653/v1/2020.acl-main.43} {A formal
  hierarchy of {RNN} architectures}.
\newblock In \emph{Proceedings of the 58th Annual Meeting of the Association
  for Computational Linguistics (ACL)}, pages 443--459.

\bibitem[{Nye et~al.(2022)Nye, Andreassen, Gur-Ari, Michalewski, Austin,
  Bieber, Dohan, Lewkowycz, Bosma, Luan, Sutton, and
  Odena}]{nye-2021-scratchpad}
Maxwell Nye, Anders Andreassen, Guy Gur-Ari, Henryk Michalewski, Jacob Austin,
  David Bieber, David Dohan, Aitor Lewkowycz, Maarten Bosma, David Luan,
  Charles Sutton, and Augustus Odena. 2022.
\newblock \href {https://openreview.net/forum?id=HBlx2idbkbq} {Show your work:
  Scratchpads for intermediate computation with language models}.
\newblock In \emph{Proceedings of the Workshop on Deep Learning for Code
  (DL4C)}.

\bibitem[{OpenAI(2023)}]{openai2023gpt4}
OpenAI. 2023.
\newblock \href {https://arxiv.org/abs/2303.08774} {{GPT}-4 technical report}.
\newblock {arXiv}:2303.08774.

\bibitem[{Paperno(2022)}]{paperno-2022-learning}
Denis Paperno. 2022.
\newblock \href {https://doi.org/10.1162/coli_a_00431} {On learning interpreted
  languages with recurrent models}.
\newblock \emph{Computational Linguistics}, 48(2):471--482.

\bibitem[{Parberry(1994)}]{parberry:1994}
Ian Parberry. 1994.
\newblock \emph{Circuit Complexity and Neural Networks}.
\newblock MIT Press.

\bibitem[{P{\'{e}}rez et~al.(2021)P{\'{e}}rez, Barcel{\'{o}}, and
  Marinkovic}]{perez-etal-2021-turing}
Jorge P{\'{e}}rez, Pablo Barcel{\'{o}}, and Javier Marinkovic. 2021.
\newblock \href {http://jmlr.org/papers/v22/20-302.html} {Attention is
  {T}uring-complete}.
\newblock \emph{Journal of Machine Learning Research}, 22:75:1--75:35.

\bibitem[{Phuong and Hutter(2022)}]{phuong2022formal}
Mary Phuong and Marcus Hutter. 2022.
\newblock \href {http://arxiv.org/abs/2207.09238} {Formal algorithms for
  transformers}.
\newblock {arXiv}:2207.09238.

\bibitem[{Pérez et~al.(2019)Pérez, Marinković, and
  Barceló}]{perez2019turing}
Jorge Pérez, Javier Marinković, and Pablo Barceló. 2019.
\newblock \href {https://openreview.net/forum?id=HyGBdo0qFm} {On the {T}uring
  completeness of modern neural network architectures}.
\newblock In \emph{Proceedings of the Seventh International Conference on
  Learning Representations (ICLR)}.

\bibitem[{Radford et~al.(2018)Radford, Narasimhan, Salimans, and
  Sutskever}]{radford2018improving}
Alec Radford, Karthik Narasimhan, Tim Salimans, and Ilya Sutskever. 2018.
\newblock \href {https://openai.com/research/language-unsupervised} {Improving
  language understanding by generative pre-training}.

\bibitem[{Reingold(2008)}]{reingold2008undirected}
Omer Reingold. 2008.
\newblock \href {https://doi.org/10.1145/1391289.1391291} {Undirected
  connectivity in log-space}.
\newblock \emph{Journal of the ACM}, 55(4):1--24.

\bibitem[{Sanford et~al.(2023)Sanford, Hsu, and
  Telgarsky}]{sanford2023representational}
Clayton Sanford, Daniel Hsu, and Matus Telgarsky. 2023.
\newblock \href
  {https://papers.nips.cc/paper_files/paper/2023/hash/73bf692447f174984f30499ec9b20e04-Abstract-Conference.html}
  {Representational strengths and limitations of transformers}.
\newblock In \emph{Advances in Neural Information Processing Systems 36
  (NeurIPS)}, pages 36677--36707.

\bibitem[{Siegelmann and Sontag(1994)}]{siegelmann+sontag:1994}
Hava~T. Siegelmann and Eduardo~D. Sontag. 1994.
\newblock \href {https://doi.org/10.1016/0304-3975(94)90178-3} {Analog
  computation via neural networks}.
\newblock \emph{Theoretical Computer Science}, 131(2):331--360.

\bibitem[{Siegelmann and Sontag(1995)}]{siegelmann+sontag:1995}
Hava~T. Siegelmann and Eduardo~D. Sontag. 1995.
\newblock \href {https://doi.org/10.1006/jcss.1995.1013} {On the computational
  power of neural nets}.
\newblock \emph{Journal of Computer and System Sciences}, 50(1):132--150.

\bibitem[{{\v S}{\'\i}ma and Orponen(2003)}]{sima+orponen:2003}
Ji{\v r}{\'\i} {\v S}{\'\i}ma and Pekka Orponen. 2003.
\newblock \href {https://doi.org/10.1162/089976603322518731} {General-purpose
  computation with neural networks: A survey of complexity theoretic results}.
\newblock \emph{Neural Computation}, 15(12):2727--2778.

\bibitem[{Sipser(2013)}]{sipser-2013-introduction}
Michael Sipser. 2013.
\newblock \emph{Introduction to the Theory of Computation}, 3rd edition.
\newblock Cengage Learning.

\bibitem[{Siu et~al.(1995)Siu, Roychowdhury, and Kailath}]{siu+:1995}
Kai-Yeung Siu, Vwani Roychowdhury, and Thomas Kailath. 1995.
\newblock \emph{Discrete Neural Computation}.
\newblock Prentice Hall.

\bibitem[{Straubing(1994)}]{straubing:1994}
Howard Straubing. 1994.
\newblock \emph{Finite Automata, Formal Logic, and Circuit Complexity}.
\newblock Springer.

\bibitem[{Strobl(2023)}]{strobl2023averagehard}
Lena Strobl. 2023.
\newblock \href {https://arxiv.org/abs/2308.03212} {Average-hard attention
  transformers are constant-depth uniform threshold circuits}.
\newblock {arXiv}:2308.03212.

\bibitem[{Sudborough(1975)}]{sudborough-1975}
I.~H. Sudborough. 1975.
\newblock \href {https://doi.org/10.1016/S0022-0000(75)80014-6} {On
  tape-bounded complexity classes and multihead finite automata}.
\newblock \emph{Journal of Computer and System Sciences}, 10(1):62–76.

\bibitem[{Suzgun et~al.(2019)Suzgun, Belinkov, Shieber, and
  Gehrmann}]{suzgun-etal-2019-lstm}
Mirac Suzgun, Yonatan Belinkov, Stuart Shieber, and Sebastian Gehrmann. 2019.
\newblock \href {https://doi.org/10.18653/v1/W19-3905} {{LSTM} networks can
  perform dynamic counting}.
\newblock In \emph{Proceedings of the Workshop on Deep Learning and Formal
  Languages: Building Bridges}, pages 44--54.

\bibitem[{Thomas(1997)}]{Thomas1997}
Wolfgang Thomas. 1997.
\newblock \href {https://doi.org/10.1007/978-3-642-59126-6_7} {Languages,
  automata, and logic}.
\newblock In Grzegorz Rozenberg and Arto Salomaa, editors, \emph{Handbook of
  Formal Languages: Volume 3 Beyond Words}, pages 389--455. Springer.

\bibitem[{Vaswani et~al.(2017)Vaswani, Shazeer, Parmar, Uszkoreit, Jones,
  Gomez, Kaiser, and Polosukhin}]{vaswani-etal-2017-attention}
Ashish Vaswani, Noam Shazeer, Niki Parmar, Jakob Uszkoreit, Llion Jones,
  Aidan~N. Gomez, Lukasz Kaiser, and Illia Polosukhin. 2017.
\newblock \href
  {https://proceedings.neurips.cc/paper/2017/hash/3f5ee243547dee91fbd053c1c4a845aa-Abstract.html}
  {Attention is all you need}.
\newblock In \emph{Advances in Neural Information Processing Systems 30
  (NeurIPS)}.

\bibitem[{Wang et~al.(2019)Wang, Li, Xiao, Zhu, Li, Wong, and
  Chao}]{wang+:2019}
Qiang Wang, Bei Li, Tong Xiao, Jingbo Zhu, Changliang Li, Derek~F. Wong, and
  Lidia~S. Chao. 2019.
\newblock \href {https://doi.org/10.18653/v1/P19-1176} {Learning deep
  {T}ransformer models for machine translation}.
\newblock In \emph{Proceedings of the 57th Annual Meeting of the Association
  for Computational Linguistics (ACL)}.

\bibitem[{Wei et~al.(2022{\natexlab{a}})Wei, Chen, and
  Ma}]{wei-etal-2021-turing}
Colin Wei, Yining Chen, and Tengyu Ma. 2022{\natexlab{a}}.
\newblock \href
  {https://papers.nips.cc/paper_files/paper/2022/hash/4ebf1d74f53ece08512a23309d58df89-Abstract-Conference.html}
  {Statistically meaningful approximation: a case study on approximating
  {T}uring machines with transformers}.
\newblock In \emph{Advances in Neural Information Processing Systems 35
  (NeurIPS)}, pages 12071--12083.

\bibitem[{Wei et~al.(2022{\natexlab{b}})Wei, Wang, Schuurmans, Bosma, Ichter,
  Xia, Chi, Le, and Zhou}]{wei-2022-cot}
Jason Wei, Xuezhi Wang, Dale Schuurmans, Maarten Bosma, Brian Ichter, Fei Xia,
  Ed~H. Chi, Quoc~V. Le, and Denny Zhou. 2022{\natexlab{b}}.
\newblock \href
  {https://proceedings.neurips.cc/paper_files/paper/2022/hash/9d5609613524ecf4f15af0f7b31abca4-Abstract-Conference.html}
  {Chain-of-thought prompting elicits reasoning in large language models}.
\newblock In \emph{Advances in Neural Information Processing Systems 35
  (NeurIPS)}, pages 24824--24837.

\bibitem[{Weiss et~al.(2018)Weiss, Goldberg, and
  Yahav}]{weiss-etal-2018-practical}
Gail Weiss, Yoav Goldberg, and Eran Yahav. 2018.
\newblock \href {https://doi.org/10.18653/v1/P18-2117} {On the practical
  computational power of finite precision {RNN}s for language recognition}.
\newblock In \emph{Proceedings of the 56th Annual Meeting of the Association
  for Computational Linguistics (ACL)}, pages 740--745.

\bibitem[{Weiss et~al.(2021)Weiss, Goldberg, and Yahav}]{weiss-etal-2021-rasp}
Gail Weiss, Yoav Goldberg, and Eran Yahav. 2021.
\newblock \href {https://proceedings.mlr.press/v139/weiss21a.html} {Thinking
  like {T}ransformers}.
\newblock In \emph{Proceedings of the 38th International Conference on Machine
  Learning (ICML)}, volume 139 of \emph{Proceedings of Machine Learning
  Research}, pages 11080--11090.

\bibitem[{Yao et~al.(2021)Yao, Peng, Papadimitriou, and
  Narasimhan}]{yao-etal-2021-self}
Shunyu Yao, Binghui Peng, Christos Papadimitriou, and Karthik Narasimhan. 2021.
\newblock \href {https://doi.org/10.18653/v1/2021.acl-long.292} {Self-attention
  networks can process bounded hierarchical languages}.
\newblock In \emph{Proceedings of the 59th Annual Meeting of the Association
  for Computational Linguistics and the 11th International Joint Conference on
  Natural Language Processing (ACL-IJCNLP)}, pages 3770--3785.

\bibitem[{Yun et~al.(2020)Yun, Bhojanapalli, Rawat, Reddi, and
  Kumar}]{yun-etal-2020-universal}
Chulhee Yun, Srinadh Bhojanapalli, Ankit~Singh Rawat, Sashank~J. Reddi, and
  Sanjiv Kumar. 2020.
\newblock \href {https://openreview.net/forum?id=ByxRM0Ntvr} {Are
  {T}ransformers universal approximators of sequence-to-sequence functions?}
\newblock In \emph{8th International Conference on Learning Representations
  ({ICLR})}.

\bibitem[{Zhou et~al.(2024)Zhou, Bradley, Littwin, Razin, Saremi, Susskind,
  Bengio, and Nakkiran}]{zhou-etal-2023-rasp-length-generalization}
Hattie Zhou, Arwen Bradley, Etai Littwin, Noam Razin, Omid Saremi, Josh
  Susskind, Samy Bengio, and Preetum Nakkiran. 2024.
\newblock \href {https://openreview.net/forum?id=AssIuHnmHX} {What algorithms
  can {T}ransformers learn? {A} study in length generalization}.
\newblock In \emph{Proceedings of the Twelfth International Conference on
  Learning Representations (ICLR)}.

\end{thebibliography}
\bibliographystyle{acl_natbib}

\end{document}